\documentclass[12pt,authoryear,review]{elsarticle}

\usepackage[bookmarks=true, colorlinks=true]{hyperref}

\usepackage{url}

\usepackage{caption}
\usepackage{amssymb,amsmath}
\usepackage{subfigure}
\usepackage{algorithm}
\usepackage{algorithmicx}
\usepackage{setspace}
\usepackage{algpseudocode}
\usepackage{multirow}
\usepackage{booktabs}
\usepackage{mathrsfs}
\usepackage{lineno}
\usepackage{multirow}\usepackage{makecell}

\journal{ISPRS Journal of Photogrammetry and Remote Sensing}

\begin{document}

\begin{frontmatter}

\title{A boundary-aware point clustering approach in Euclidean and embedding spaces for roof plane segmentation }
\author[1,3]{Li Li } 
\author[2]{Qingqing Li } 
\author[1]{Guozheng Xu } 
\author[1]{Pengwei Zhou } 
\author[2]{Jingmin Tu }
\author[2]{Jie Li}
\author[4]{Mingming Li}
\author[1,3]{Jian Yao\corref{CorAuthor}}

\cortext[CorAuthor]{Corresponding author.}
\ead{jian.yao@whu.edu.cn}	
\address[1]{School of Remote Sensing and Information Engineering, Wuhan University, Wuhan
 430079, P.R. China}
\address[2]{School of Electrical and Electronic Engineering, Hubei University of Technology, Wuhan, Hubei, China}
\address[3]{Wuhan University Shenzhen Research Institute, Shenzhen 518057, China.}
\address[4]{Soarscape Technology (Shanghai) Co., Ltd.}

\begin{abstract}
Roof plane segmentation from airborne light detection and ranging (LiDAR) point clouds is an important technology for three-dimensional (3D) building model reconstruction. One of the key issues of plane segmentation is how to design powerful features that can exactly distinguish adjacent planar patches. The quality of point feature directly determines the accuracy of roof plane segmentation. Most of existing approaches use handcrafted features, such as point-to-plane distance, normal vector, etc., to extract roof planes. However, the abilities of these features are relatively low, especially in boundary areas. To solve this problem, we propose a boundary-aware point clustering approach in Euclidean and embedding spaces constructed by a multi-task deep network for roof plane segmentation. We design a three-branch multi-task network to predict semantic labels, point offsets and extract deep embedding features. In the first branch, we classify the input data as non-roof, boundary and plane points. In the second branch, we predict point offsets for shifting each point towards its respective instance center. In the third branch, we constrain that points of the same plane instance should have the similar embeddings. We aim to ensure that points of the same plane instance are close as much as possible in both Euclidean and embedding spaces. However, although deep network has strong feature representative ability, it is still hard to accurately distinguish points near the plane instance boundary. Therefore, we first robustly group plane points into many clusters in Euclidean and embedding spaces to find candidate planes. Then, we assign the rest boundary points to their closest clusters to generate the final complete roof planes. In this way, we can effectively reduce the influence of unreliable boundary points. In addition, to train the network and evaluate the performance of our approach, we prepare a synthetic dataset and two real datasets. The experiments conducted on synthetic and real datasets show that the proposed approach significantly outperforms the existing state-of-the-art approaches in both qualitative evaluation and quantitative metrics. To facilitate future research, we will make datasets and source code of our approach publicly available at \url{https://github.com/Li-Li-Whu/DeepRoofPlane}.

\end{abstract}

\begin{keyword}
Roof plane segmentation \sep building reconstruction \sep airborne LiDAR \sep point clouds \sep deep learning 
\end{keyword}

\end{frontmatter}
%


\section{Introduction}
\label{sec:introduction}
Recently, three-dimensional (3D) building models at different Levels of Detail (LoDs) have attracted much attention in Photogrammetry and computer vision fields, and have been widely applied in many urban applications, such as smart cities, urban planning, property management and navigation systems. Airborne light detection and ranging (LiDAR) data is one of the widely used data sources for building segmentation~\citep{liu2024building} and 3D building model reconstruction~\citep{wang2018lidar}, because it can quickly acquire the 3D information of urban areas. To automatically reconstruct 3D building models from airborne LiDAR point clouds, one of the key technologies is the roof plane segmentation~\citep{sampath2009segmentation, xu2015investigation, fang2018planar, li2020roof, xu2020plane, zhang2022improved, liu2023roof}. However, it still remains an open and challenging issue for existing building reconstruction systems~\citep{lafarge2011building, xiong2015flexible, nan2017polyfit, liu2019topolap, xie2021combined, liu2023generation} how to exactly segment roof planes from input 3D point clouds. 

As we know, the results of roof plane segmentation highly depend on the quality of point features. Most of existing roof plane segmentation approaches design some handcrafted features, such as point-to-plane distance and normal vector, to describe points at first. Then, these approaches use region growing~\citep{vo2015octree, liu2023roof}, clustering~\citep{chen2023local}, or RANSAC~\citep{canaz2020improved, li2022ransac} method to extract roof planes based on the specifically designed features. However, these approaches may generate roof planes with low quality. For example, region growing and clustering may fail to find accurate boundaries between adjacent roof planes, and may cause the problems of over- and under-segmentation. RANSAC-based approaches may detect spurious planes. The key reason is that the representation abilities of handcrafted features they used are relatively low, especially in boundary areas. To improve the performance of these approaches, some energy-based methods, such as boundary relabeling~\citep{li2020roof,liu2023roof} and global energy optimization~\citep{yan2014global, dong2018efficient, wang2021roof}, are proposed to further refine their results. However, these optimization-based approaches also use some handcrafted prior knowledge to construct energy cost functions.

In recent years, because deep networks can automatically learn powerful features from the input large-scale data, deep learning-based methods have shown excellent performance in many tasks of point cloud processing~\citep{guo2020deep}. Recently, some deep networks are also designed for plane recovery~\citep{liu2018planenet, liu2019planercnn, tan2021planetr, liu2022planemvs}. However, these networks are designed for plane recovery from two-dimensional (2D) images, and may be unsuitable for roof plane segmentation from 3D point clouds. Similar to the method presented in ~\citep{zhang2022improved}, we also attempt to propose a deep learning-based roof plane segmentation approach. 
~\cite{zhang2022improved} first design a network named RoofNet based on ASIS~\citep{wang2019associatively} to learn pointwise instance embeddings, then a clustering algorithm is applied to group points into planar clusters. 
However, there are two issues that can be improved. First, the deep network attempts to separate plane instances in embedding space, but ignores that the Euclidean distances between points of adjacent plane instances are still close. Namely, it ignores the spatial distributions of plane instances. Either embedding or Euclidean space may be not powerful enough to distinguish adjacent roof planes.
Second, although deep networks have stronger feature representation ability than traditional handcrafted methods, they still struggle to extract accurate features for points near the plane instance boundary. The boundaries of segmented planes may be not smooth if we treat all points the same. 

To solve the above-mentioned issues, we propose a boundary-aware point clustering approach in Euclidean and embedding spaces constructed by deep network for roof plane segmentation from airborne LiDAR data. The key ideas of our approach are: (1) We aim to ensure that points of the same roof plane instance are close as much as possible in both Euclidean and embedding spaces, so we can accurately divide points into many planar patch clusters; (2) We try to identify boundary points at first and then process them individually, so we can reduce the influence of unreliable boundary points. To meet our requirements, inspired by the existing 3D point cloud instance segmentation networks~\citep{pham2019jsis3d, jiang2020pointgroup, vu2022softgroup}, we design a three-branch network to identify boundary points and construct Euclidean and embedding spaces for point clustering. We first group plane points into many planar clusters using the two spaces, and we further assign each boundary point to its closest cluster center. In addition, we construct a synthetic and two real datasets to train and evaluate our approach. The key contributions of our approach are summarized as follows:
\begin{itemize}
\item Inspired by 3D point cloud instance segmentation networks, we design a three-branch network to simultaneously construct the Euclidean and embedding spaces for plane point clustering. 
\item Instead of treating all points the same, we propose to identify boundary and plane points at first and then process them using different strategies. Thus, we can avoid the influence of unreliable boundary points as much as possible.
\item We construct a synthetic dataset and two real datasets for roof plane segmentation network training and performance evaluation. We will make our datasets publicly available to facilitate future research.
\end{itemize}
%

\section{Related work}
\label{sec:relatedwork}

In recent decades, many roof plane segmentation approaches have been proposed. Generally, we can divide these approaches into three categories: region growing-based, model fitting-based, energy-based, feature clustering-based, and deep learning-based approaches. 

\subsection{Region growing-based approaches}
\label{sec::rgmethods}
Region growing technology is simple but effective, and has been widely applied for roof plane segmentation. Region growing-based approaches~\citep{gorte2002segmentation, vo2015octree, cao2017roof, xu2017segmentation, gilani2018segmentation, luo2021supervoxel} first select the seeds from input points, and then expand to adjacent points using some specially designed criteria. For example, ~\cite{vo2015octree} proposed a coarse-to-fine region growing-based plane segmentation method. This method first performs region growing at voxel level to extract coarse planar patches. Then, unlabeled points are assigned to the closest planes. ~\cite{cao2017roof} proposed to select robust seed points in a parameter space at first. Then, the region growing is performed in the spatial space to segment roof planes.~\cite{gilani2018segmentation} first calculated point normal vectors and detected boundary points using an improved Principal Component Analysis (PCA) method. Then, a non-boundary point with the minimal curvature value is selected as the seed. Last, region growing is performed to extract roof planes. To reduce the influence of noises, ~\cite{luo2021supervoxel} proposed to perform region growing at supervoxel-level instead of point-level. However, region growing-based methods have two essential disadvantages. First, the roof plane segmentation results are sensitive to the selection of seeds. Second, region growing-based methods are hard to find the accurate boundaries between two adjacent planar patches, especially for smooth regions. Thus, the problems of jagged boundaries, over- and under-segmentation may appear. 

\subsection{Model fitting-based approaches}
\label{sec::mfmethods}

Model fitting-based approaches mainly use Hough Transform (HT) and Random Sample Consensus (RANSAC) to robustly find points that satisfy plane model. HT-based approaches~\citep{overby2004automatic, borrmann20113d, hulik2014continuous} first transform the input points into the parameter space, and then find planes by a voting procedure. ~\cite{tarsha2007hough} compared the performance of HT and RANSAC in roof plane segmentation, and concluded that RANSAC is a better choice for roof plane segmentation because RANSAC has a low computational cost and is more robust than HT. RANSAC-based approaches~\citep{schnabel2007efficient, chen2012urban, xu2015investigation, li2017improved, dal2020adaptive, canaz2020improved, li2022ransac} first randomly sample several points to generate a large number of candidate planar patches, and then select the best one as the final planar patch. Although RANSAC-based approaches can robustly extract planar patches from noised points, they often suffer from the problem of spurious planes. Numerous variants have been proposed to alleviate this problem. For example, ~\cite{xu2015investigation} proposed an improved weighted RANSAC algorithm to segment roof planes. The weight functions they used are defined by considering both point-to-plane distance and normal vector difference.~\cite{li2017improved} proposed to improve the standard RANSAC algorithm using the Normal Distribution Transformation (NDT) cell. Instead of taking point as the sample unit, this method regards NDT cell as a minimal sample to improve the quality of sampling.~\cite{dal2020adaptive} proposed an adaptive RANSAC algorithm for roof segmentation. In their method, the distance and angle thresholds are adaptively determined to check the consistency between points and the hypothetical plane model. However, the spurious plane problem is still not completely solved in RANSAC-based approaches. 

\subsection{Energy-based approaches}
\label{sec::enermethods}
To solve the problems that existed in region-growing or model fitting-based approaches, many energy-based methods~\citep{yan2014global, dong2018efficient, li2020roof, wang2021roof, yu2022finding, liu2023roof} have been proposed to further refine the initial planar patches. These approaches usually formulate plane refinement as an energy function optimization problem.~\cite{yan2014global} first detected initial roof planes using region growing. Then, they formulated plane refinement as a multi-label optimization problem~\citep{isack2012energy}, and globally solved it using graph cuts~\citep{boykov2001fast}.~\cite{dong2018efficient} and ~\cite{wang2021roof} also applied graph cuts to refine the initial roof planar patches. However, the computational cost of multi-label optimization is high. To improve the efficiency of roof plane refinement, some methods applied boundary relabeling~\citep{li2020roof,liu2023roof,li2023roof} to gradually refine the boundaries of initial roof planar patches.~\cite{yu2022finding} first designed a multi-object energy function that considers high fidelity, high simplicity and high completeness altogether. Then, they proposed a novel iterative optimization mechanism to refine the initial planar patches by minimizing the energy function. However, this method may converge to the local optimum if the quality of initial planar patches is bad. Although these energy-based approaches perform well in many cases, they still rely on handcrafted prior knowledge or features to design energy cost functions.

\subsection{Feature clustering-based approaches}
\label{sec::fcmethods}
Feature clustering-based approaches~\citep{filin2006segmentation, sampath2009segmentation, zhang2018spectral, xu2019plane, li2021novel, zhang2022improved, chen2023local} use some point clustering methods to group points into many planar patches in the feature space.~\cite{filin2006segmentation} proposed a slope adaptive neighborhood system to calculate accurate and reliable point attributes. Then, the points are segmented by clustering them in the attribute space. ~\cite{sampath2009segmentation} applied the fuzzy k-means algorithm to cluster points into planar patches based on the point normals.~\cite{zhang2018spectral} proposed to apply the spectral clustering method to detect roof planes from airborne LiDAR data. This approach groups straight-line segments instead of points.~\cite{chen2023local} first designed a new measurement metric based on the tangent plane distance to evaluate the similarities of the input points. Then, an adaptive DBSCAN is applied to cluster coplanar points. The key issue of the feature clustering-based approaches is how to design the appropriate features that can effectively distinguish different plane instances. However, because the representative abilities of handcrafted features are relatively low, some problems, such as inaccurate boundary and over-segmentation, may still appear.

\subsection{Deep learning-based approaches}
\label{sec::dpmethods}

Recently, some deep learning-based instance segmentation~\citep{pham2019jsis3d, jiang2020pointgroup, he2020instance, vu2022softgroup, he2022dynamic, zhao2023divide} and primitive segmentation approaches~\citep{yan2021hpnet, huang2021primitivenet} have been proposed. In general, these methods first apply deep networks trained by metric loss functions to extract discriminative features, then point clustering or region growing methods are applied to extract instances or primitives. HPNet~\citep{yan2021hpnet} combined three learned features with adaptive weights and applied mean-shift clustering to extract primitive instances. PrimitiveNet~\citep{huang2021primitivenet} designed an embedding sub-network to extract high-dimensional local features and a discriminator sub-network to determine whether two features belong to the same primitive instance. The outputs of these two sub-networks are further used for primitive instance segmentation. However, these networks are not designed for roof plane segmentation.~\cite{zhang2022improved} proposed to use deep features for roof plane clustering. This method first designs a multi-task deep network to map point clouds into a high-dimensional embedding space. It attempts to ensure that points of the same roof plane have the similar embedding features. Then, a point clustering method is performed to extract roof planes from the embedding space. However, this method may also fail in the boundary regions because deep network also struggles to extract accurate embedding features for points near boundaries.

\section{Our approach}
\label{sec:approach}
\begin{figure}[!htb]
	\centering
	{\includegraphics[width= 0.99 \linewidth]{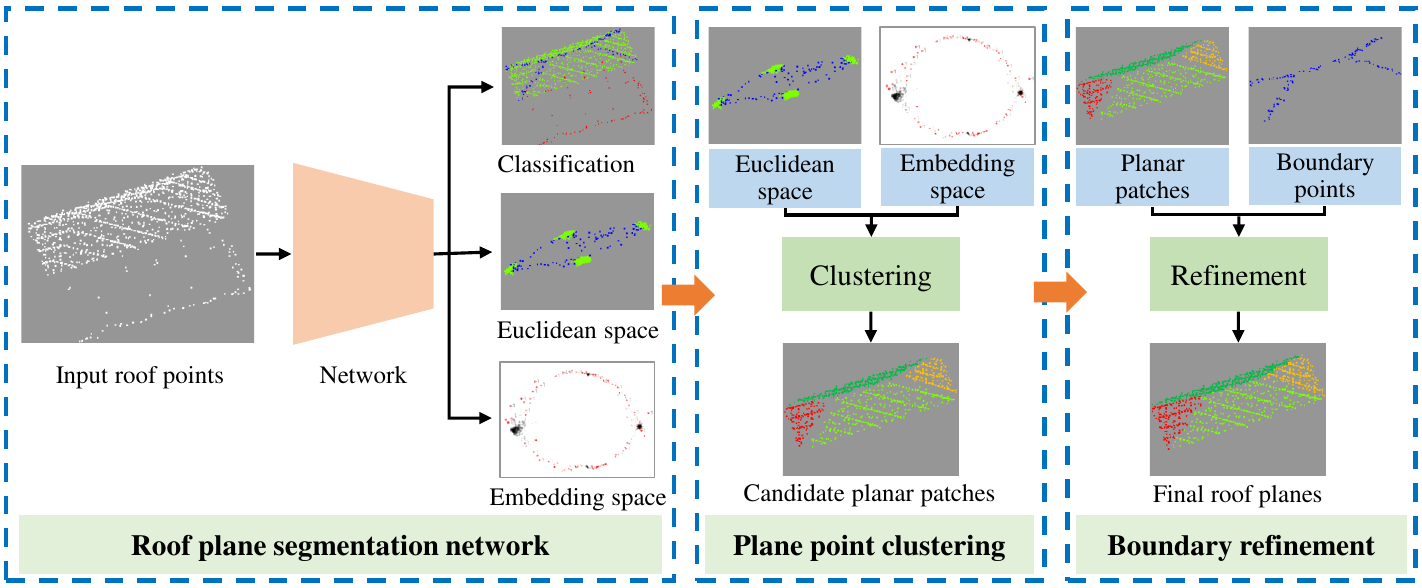}} 
	\caption{The whole framework of the proposed roof plane segmentation approach.}
	\vspace{0.0em}
	\label{Fig:workflow}
\end{figure}

Let $\mathbf{P} = \{\mathbf{p}_i\}_{i=0}^N$ denote the input roof points of our approach, where $N$ is the point number. The goal of roof plane segmentation approach is to accurately segment $\mathbf{P}$ into many planar patches. To achieve this goal, we propose a boundary-aware point clustering approach in Euclidean and embedding spaces constructed by deep network. As shown in Figure~\ref{Fig:workflow}, the proposed approach consists of three steps. In the first step, we design a three-branch multi-task roof plane segmentation network to classify the input points and construct attribute spaces for point clustering. In the second step, we cluster plane points into many candidate planar patches using Euclidean and embedding spaces constructed by the deep network. In the last step, to generate the final roof planes, the unlabeled boundary points are assigned to their closest planar patches. 

%

\subsection{Roof plane segmentation network}
\label{sec:deepnetwork}
\begin{figure}[!htb]
	\centering
	{\includegraphics[width= 0.99 \linewidth]{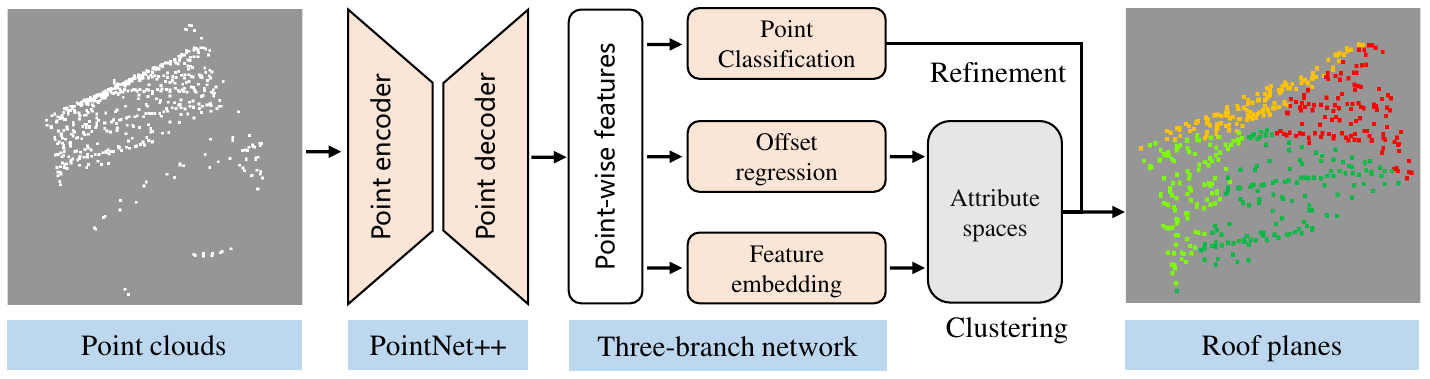}} 
	\caption{The brief architecture of the proposed roof plane segmentation network.}
	\vspace{0.0em}
	\label{Fig:network}
\end{figure}

In this paper, we formulate roof plane segmentation as an instance segmentation problem. We design a three-branch multi-task network to solve roof plane instance segmentation problem. In the first branch, we classify the input $\mathbf{P}$ as non-roof, boundary and plane points. Different strategies are designed to process different classes of point clouds. In the second and third branches, we simultaneously construct the Euclidean and embedding spaces for roof point clustering. We attempt to ensure that points of the same plane instance are close as much as possible in both Euclidean and embedding spaces. In this way, we can easily group points into different planar patches. The brief network architecture is shown in Figure~\ref{Fig:network}.

%
%

\subsubsection{Network architecture}
\label{sec:architecture}

In the proposed roof segmentation network, we first extract point-wise features for the input roof points $\mathbf{P}$ using PointNet++ network~\citep{qi2017pointnet2}. It should be noted that other networks can also be applied to extract point-wise features in our approach. Then, we perform point classification and attribute space construction based on the extracted point-wise features. 

\begin{figure}[!htb]
	\scriptsize
	\centering
	\begin{tabular}{cccc}
		\multicolumn{1}{c}{\includegraphics[width= 0.225 \linewidth]{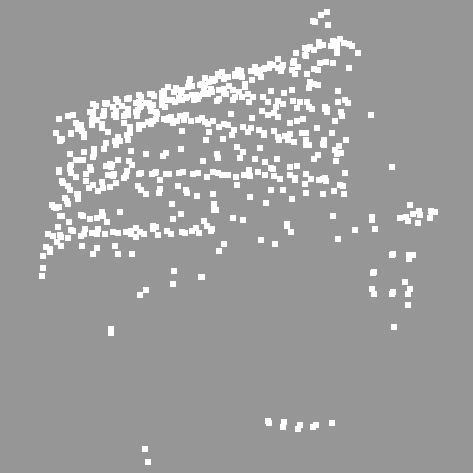}} &
		\multicolumn{1}{c}{\includegraphics[width= 0.225 \linewidth]{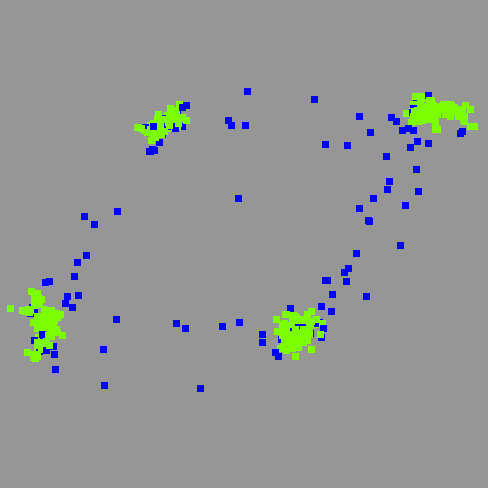}} &
		\multicolumn{1}{c}{\includegraphics[width= 0.225 \linewidth]{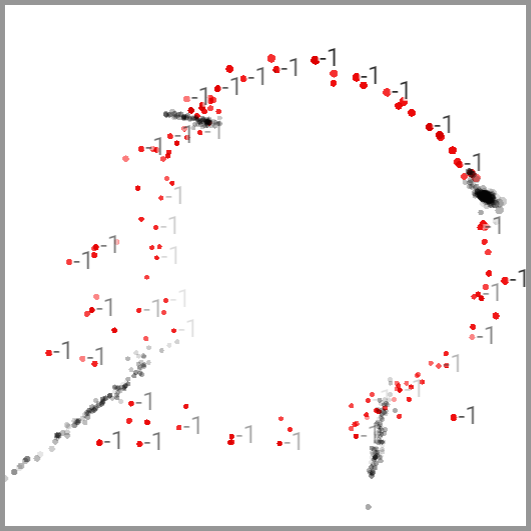}} &
		\multicolumn{1}{c}{\includegraphics[width= 0.225 \linewidth]{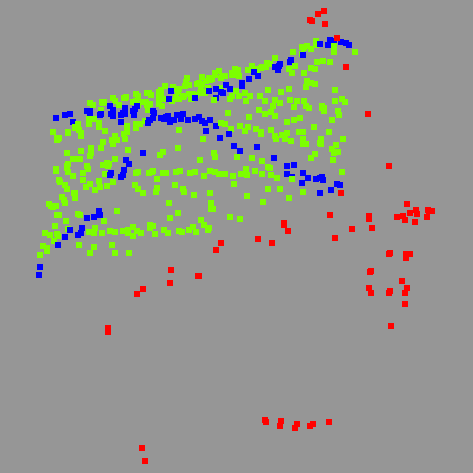}} \\
		\multicolumn{1}{c}{(a)} & \multicolumn{1}{c}{(b)} & \multicolumn{1}{c}{(c)} & \multicolumn{1}{c}{(d)}
	\end{tabular} 
	\caption{Illustration of the point classification branch. This figure presents the (a) input points with roof and non-roof points, the (b) shifted points, the (c) embedding features of points and the (d) final point classification result. In (a), (b) and (d), the white, green, blue and red dots represent the input, roof plane, boundary and non-roof points. In (c), the red dots labeled with $-1$ represent boundary points, other gray dots represent plane points.}
	\vspace{0.0em}
	\label{Fig:class}
\end{figure}

In the first branch, we design a point classification head to classify the input $\mathbf{P}$ as non-roof, boundary and plane points. In real applications, the input points inevitably contain some non-roof points, such as noise, facade and tree points. Obviously, these non-roof points will influence the results of roof plane segmentation. Thus, we need to classify non-roof points and remove them from the input $\mathbf{P}$. In addition,  because deep network still struggles to extract accurate features for boundary points, we also need to find boundary points from $\mathbf{P}$ before performing point clustering. In Figure~\ref{Fig:class}, we present an example to illustrate why the point classification is necessary. In Figure~\ref{Fig:class}(a), we present the input points and we can find that there are many non-roof points. In Figure~\ref{Fig:class}(b) and (c), we present the shifted roof points (Euclidean space) and the point-wise embedding features (embedding space). We visualize the high-dimensional embedding features using Embedding Projector~\citep{smilkov2016embedding}. We observe that many boundary points (blue points in (b) and red points in (c)) are not close to the instance centers in both spaces. Namely, the features of boundary points are unreliable. This is the main reason that we need to classify boundary points from the input point clouds before performing point clustering. We present the final point classification result in Figure~\ref{Fig:class}(d), and we can find that the proposed network can accurately classify the input points into three categories. Let $\mathbf{P}_{n}$, $\mathbf{P}_b$ and $\mathbf{P}_r$ denote non-roof, boundary and plane points. 

In the second and third branches, we design an offset regression head and a feature embedding head to simultaneously construct Euclidean and embedding spaces for point clustering. From Figure~\ref{Fig:class}(b) and (c), we find that either Euclidean and embedding spaces cannot completely distinguish different roof plane instances. Thus, we attempt to combine these two spaces to facilitate point clustering. In the offset regression head, for each point $\mathbf{p}_i$, we apply a MLP (Multilayer Perception) layer to predict 3D offset $\Delta \mathbf{p}_i$ from $\mathbf{p}_i$ to the associated roof plane instance center. Then, we shift points into the instance centers. In this way, we can distinguish different roof plane instances in the Euclidean space. In the feature embedding head, we transform point-wise features to the high-dimensional embedding space using a MLP layer. We constrain that points of the same plane instance have the similar embedding features. 

\subsubsection{Label generation}
\label{sec:Label}
%
In the input points $\mathbf{P}$, there are two kinds of ground truth labels. The first kind is the ground truth of instance segmentation. This label indicates the plane instance that each point belongs to. The second kind indicates whether a point is non-roof point. However, these two kinds of labels are not enough to train the proposed network. We also need to automatically find boundary points from the input $\mathbf{P}$ and calculate offset vector for each point included in $\mathbf{P}$.

For each plane point $\mathbf{p}_i$, we first find the $k$ nearest points $\mathcal{N}(\mathbf{p}_i)$. If there is more than one point in $\mathcal{N}(\mathbf{p}_i)$ that belongs to the different plane instance, we label point $\mathbf{p}_i$ as a boundary point. In addition, the ground truth offset vector $\Delta_{gt}(\mathbf{p}_i)$ is calculated as $\Delta_{gt}(\mathbf{p}_i) = ||\hat{\mathbf{c}}_i-\mathbf{p}_i||$, where $\hat{\mathbf{c}}_i$ represents the center of the plane instance that $\mathbf{p}_i$ belongs to.

\subsubsection{Loss functions}
\label{sec:loss}

The loss functions $\mathcal{L}$ of the proposed roof plane segmentation network consist of the point classification loss $\mathcal{L}_{cls}$, the offset regression loss $\mathcal{L}_{reg}$ and the embedding loss $\mathcal{L}_{emb}$. We calculate $\mathcal{L}$ as:
\begin{equation}
	\mathcal{L} = \mathcal{L}_{cls} + \mathcal{L}_{reg} + \mathcal{L}_{emb}.
\end{equation}

We use the classical cross-entropy loss to define the point classification loss $\mathcal{L}_{cls}$. $\mathcal{L}_{cls}$ is calculated as:
\begin{equation}
	\mathcal{L}_{cls} = \sum_{\mathbf{p}_i\in\mathbf{P}} \mathcal{L}_{ce}(l_{cls}^{prd}(\mathbf{p}_i), l_{cls}^{gt}(\mathbf{p}_i)),
\end{equation}
where $\mathcal{L}_{ce}(\cdot)$ calculates cross-entropy loss value. $l_{cls}^{prd}(\mathbf{p}_i)$ and $l_{cls}^{gt}(\mathbf{p}_i)$ are the predicted classification label and the ground truth label for point $\mathbf{p}_i$, respectively.

The offset regression loss $\mathcal{L}_{reg}$ is applied to constrain that points of the same instance are close in Euclidean space. $\mathcal{L}_{reg}$ is defined similar to PointGroup~\citep{jiang2020pointgroup} as follows:
\begin{equation}
	\mathcal{L}_{reg} = \sum_{\mathbf{p}_i \in \mathbf{P}} (||\Delta \mathbf{p}_i - \Delta_{gt}(\mathbf{p}_i)|| + \frac{\Delta \mathbf{p}_i}{|\Delta \mathbf{p}_i|} \cdot \frac{\Delta_{gt}\mathbf{p}_i}{|\Delta_{gt}(\mathbf{p}_i)|} ),
\end{equation}
where $\Delta \mathbf{p}_i$ and $\Delta_{gt}\mathbf{p}_i$ are the predicted and ground truth offset vectors, respectively. $\mathcal{L}_{reg}$ aims to constrain that the values and directions of the predicted offset vectors are the same with the corresponding ground truth.

The embedding loss $\mathcal{L}_{emb}$ is applied to constrain that the point embeddings belonging to the same plane instance are similar. 
In this paper, we directly use the same embedding loss presented in  JSIS3D~\citep{pham2019jsis3d} and ASIS~\citep{wang2019associatively}. The embedding loss is defined as:
\begin{equation}
	\mathcal{L}_{emb} = \mathcal{L}_{pull} + \mathcal{L}_{push} + 0.001 \mathcal{L}_{regularization},
\end{equation}
where $\mathcal{L}_{pull}$ aims to pull embeddings towards the instance centers, $\mathcal{L}_{push}$ attempts to push instance centers away from each other, and $\mathcal{L}_{regularization}$ is applied to draw centers towards the origin. For the detailed descriptions of $\mathcal{L}_{pull}$, $\mathcal{L}_{push}$ and $\mathcal{L}_{regularization}$, please refer to JSIS3D~\citep{pham2019jsis3d} or ASIS~\citep{wang2019associatively}.

\subsection{Plane point clustering}
\label{sec:clustering}

To automatically generate planar patches, inspired by PointGroup~\citep{jiang2020pointgroup},  we need to group points into many clusters. In our approach, we use both Euclidean and embedding spaces to perform point clustering. To eliminate the influence of non-roof points $\mathbf{P}_{n}$, we directly remove them before performing point clustering. Although deep network has powerful feature representative ability, it still struggles to extract accurate features for points near instance boundary. Thus, to reduce the influence of unreliable boundary points $\mathbf{P}_{b}$, we only perform point clustering for plane points $\mathbf{P}_{r}$. The detailed description of point clustering method is presented in Algorithm~\ref{Alg:pointclustering}.
\begin{algorithm}[!htb]
	\setstretch{1.0}
	\caption{Roof plane point clustering method.}
	\label{Alg:pointclustering}
	\renewcommand{\algorithmicrequire}{\textbf{Input:}}
	\renewcommand{\algorithmicensure}{\textbf{Output:}}
	\begin{algorithmic}[1]
		\Require Clustering radius $r$; Shifted roof plane points $\mathbf{P}_r^\prime$; Embedding features $\mathbf{F}_r$ of roof plane points.  \vspace{2.5 pt}
		\Ensure  Planar clusters $\mathbf{C} = \{C_1, \cdots , C_M\}$.  \vspace{2.5 pt}
		\\ \textbf{Initialization}: An empty cluster $\mathbf{C}$; A mask array $\mathbf{A} = \{A_i, \cdots, A_{N_r}\}$ with all zero.
		\\ \textbf{for} $i = 0$ to $N_r$ \textbf{do} 
		\\  \qquad \textbf{if} $A_i == 0$ \textbf{then}
		\\  \qquad \qquad set $A_i = 1$; create an empty queue $Q$; create an empty cluster $C$
		\\  \qquad \qquad insert $i$ to $Q$; push $i$ to $C$
		\\  \qquad \qquad \textbf{while} $Q$ is not empty \textbf{do}
		\\  \qquad \qquad \qquad k = Q.getFirst() 
		\\  \qquad \qquad \qquad remove $k$ from $Q$
		\\  \qquad \qquad \qquad \textbf{for} j = 0 to $N_r$ \textbf{do}
		\\  \qquad \qquad \qquad \qquad \textbf{if} $A_j == 1$ \textbf{then}
		\\  \qquad \qquad \qquad \qquad \qquad continue
		\\  \qquad \qquad \qquad \qquad \textbf{else}
		\\  \qquad \qquad \qquad \qquad \qquad calculate $d_e = || \mathbf{p}_j - \mathbf{p}_k ||_2$ and $d_f = || \mathbf{f}_j - \mathbf{f}_k ||_2$
		\\  \qquad \qquad \qquad \qquad \qquad \textbf{if} $(w_1 d_e + w_2 d_f) < r$ \textbf{then}
		\\  \qquad \qquad \qquad \qquad \qquad \qquad set $A_j = 1$; insert $j$ to $Q$;  push $j$ to $C$
		\\  \qquad \qquad \qquad \qquad \qquad \textbf{end if}
		\\  \qquad \qquad \qquad \qquad \textbf{end if}
		\\  \qquad \qquad \qquad \textbf{end for}
		\\  \qquad \qquad \textbf{end while}
		\\  \qquad \qquad \textbf{if}  $|C| > T_n$, \textbf{then}
		\\  \qquad \qquad \qquad push $C$ to $\mathbf{C}$
		\\  \qquad \qquad \textbf{else}
		\\  \qquad \qquad \qquad continue
		\\  \qquad \qquad \textbf{end if}
		\\  \qquad \textbf{end if}
		\\ \textbf{end for}
		\\ \textbf{return} $\mathbf{C}$.
	\end{algorithmic}
\end{algorithm} 

In Algorithm~\ref{Alg:pointclustering}, $N_r$ is the number of roof plane points. $M$ is the number of clusters. $w_1$ and $w_2$ are two weights to balance distance calculated in Euclidean and embedding spaces. In general, we set $w_1 = 0.1$ and $w_2 = 0.9$. $|C|$ denotes the number of points in cluster $C$. $T_n$ is a predefined threshold that is used to remove the clusters with small elements, and we set $T_n = 50$ in our experiments. Each cluster in $\mathbf{C}$ represents a planar patch. The core idea of our point clustering approach is that for each paired points $(\mathbf{p}_i, \mathbf{p}_j)$, we evaluate their similarity using both Euclidean distance and embedding feature similarity. It should be noted that other point clustering methods also can be applied to group roof plane points in our approach.

\subsection{Boundary point refinement}
\label{sec:boudary}

To generate final roof planes, we need to further assign a label for each unlabeled point. It should be noted that most of unlabeled points in our approach are boundary points, but they may also include a few unlabeled plane points. For convenience of illustration, we call all unlabeled points as boundary points. Next, we will introduce a simple but effective boundary point refinement method. 

For each planar patch $C_k \in \mathbf{C}$, we estimate the plane parameters of $C_k$ using the included points. We also calculate the embedding center feature $\bar{\mathbf{f}_{k}}$ of $C_k$ by averaging all embedding features of points included in $C_k$. Then, for each boundary point $\mathbf{p}_i$, we calculate its distance to all planar patches. The distance $d$ between $\mathbf{p}_i$ and $C_k$ is defined as:
\begin{equation}
	d = D(\mathbf{p}_i, C_k) + ||\mathbf{f}_i - \bar{\mathbf{f}_{k}}||_2
\end{equation}
where $D(\cdot)$ means point-to-plane distance. For each point $\mathbf{p}_i$, we directly assign it to the cluster with the minimum distance $d$. Let $\widetilde{\mathbf{C}} = {\widetilde{C}_1, \cdots, \widetilde{C}_M}$ denote the final refined planar patches.

\begin{figure}[!htb]
	\centering
	\begin{tabular}{cccc}
		\multicolumn{1}{c}{\includegraphics[width= 0.225 \linewidth]{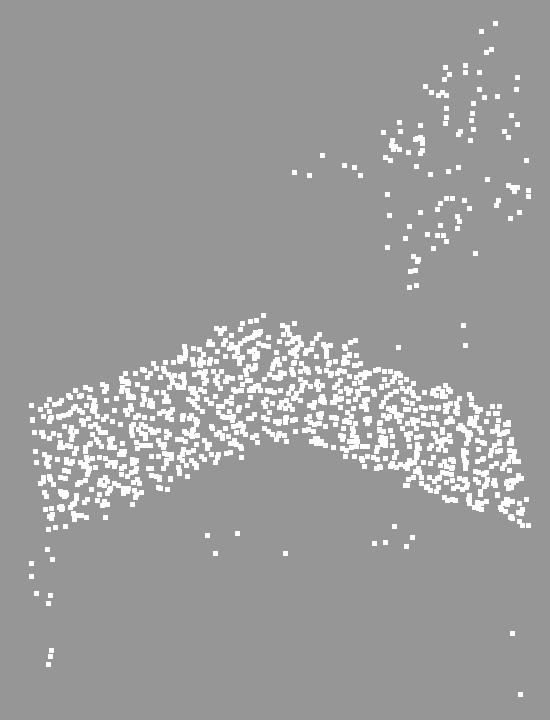}} &
		\multicolumn{1}{c}{\includegraphics[width= 0.225 \linewidth]{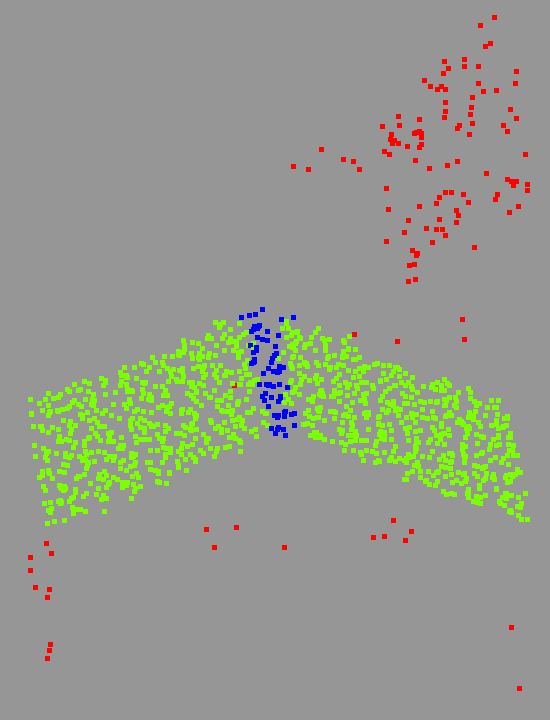}} &
		\multicolumn{1}{c}{\includegraphics[width= 0.225 \linewidth]{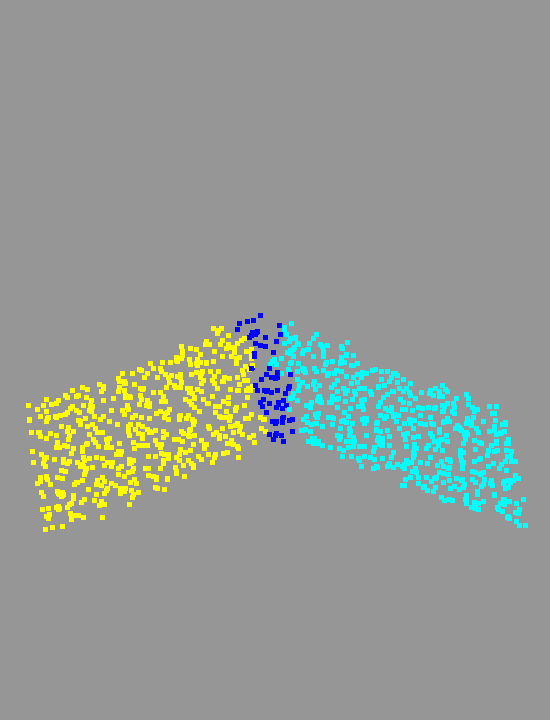}} &
		\multicolumn{1}{c}{\includegraphics[width= 0.225 \linewidth]{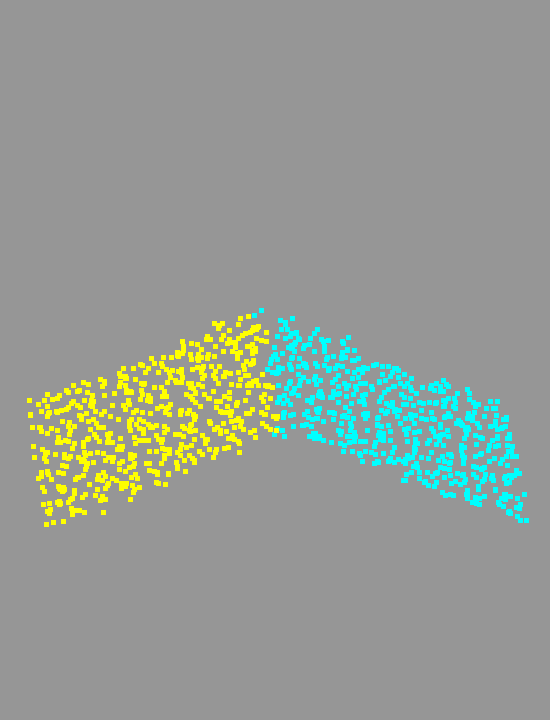}} \\
		\multicolumn{1}{c}{(a)} & \multicolumn{1}{c}{(b)} & \multicolumn{1}{c}{(c)} & \multicolumn{1}{c}{(d)}
	\end{tabular} 
	\caption{An example of plane point clustering and boundary point refinement. (a) is the input point clouds with two roof planes. (b) presents the result of point classification, where red, blue and green colors represent non-roof, boundary and plane points. (c) shows boundary points and the roof planar patches generated by plane point clustering . (d) is the final roof plane segmentation result. In (c) and (d), we apply yellow and cyan dots to denote two planar patches.}
	\vspace{0.0em}
	\label{Fig:plane}
\end{figure}

To better illustrate the proposed boundary-aware point clustering approach, we present an example in Figure~\ref{Fig:plane}. In Figure~\ref{Fig:plane}(a), we present the input point clouds and we can find that there are two adjacent roof planes in this building. The input points are classified as non-roof, boundary and plane points, as shown in Figure~\ref{Fig:plane}(b). The plane points are accurately grouped into two planar patches, as shown in Figure~\ref{Fig:plane}(c). After boundary refinement, the boundary points are also accurately assigned to the planar patches, as shown in Figure~\ref{Fig:plane}(d).

%

\section{Experimental results}
\label{sec:experiments}
%
\subsection{ Implementation details}
\label{sec:detail}

\textbf{Network architecture details.} The network architecture used in our approach is simple but effective. We select PointNet++~\citep{qi2017pointnet2} as the backbone to extract point-wise features. The number of the input points is $2048$ and the channel is $3$. We apply four Multi-scale Set Abstraction (MSA) modules to encode the input points. The multi-scale receptive radii are set as $(0.05, 0.1)$, $(0.1, 0.2)$, $(0.2, 0.4)$, and $(0.4, 0.8)$. We downsample the input points to $1024$, $256$, $64$, and $16$ in four MSA modules.
Then, we apply four Feature
Propagation (FP) modules to decode features back to $2048$ points with $128$-dimensional features. Namely, the size of the final point-wise features is $2048 \times 128$.

The point classification, offset regression, and feature embedding heads have similar structures. Based on the $128$-dimensional point-wise features, we directly apply three MLP layers to produce semantic classification labels, offset vectors, and $64$-dimensional embedding features, respectively.

\textbf{Training details.} We apply ADAM~\citep{kingma2014adam} optimizer to train the proposed roof plane segmentation network. The learning rate, weight decay, batch size, maximum number of epochs are set to $1 \times 10^{-3}$, $1 \times 10^{-3}$, $16$, and $100$, respectively. We decrease the learning rate every 20 epochs by multiplying $0.5$. We also set that the learning rate of each epoch should be larger than $1\times 10^{-6}$. 

\subsection{ Dataset preparation}
\label{sec:dataset}

\begin{figure}[!htb]
	\centering
	\begin{tabular}{cccc}
		\multicolumn{1}{c}{\includegraphics[width= 0.225 \linewidth]{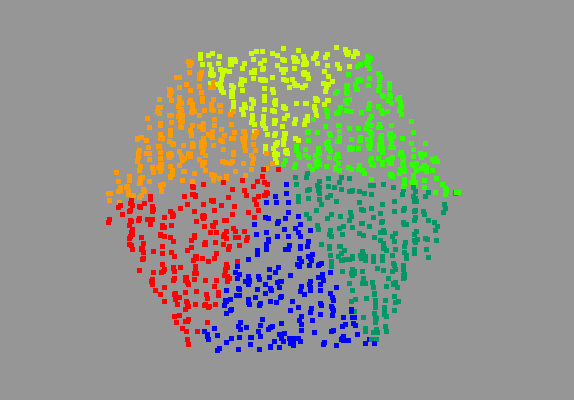}} &
		\multicolumn{1}{c}{\includegraphics[width= 0.225 \linewidth]{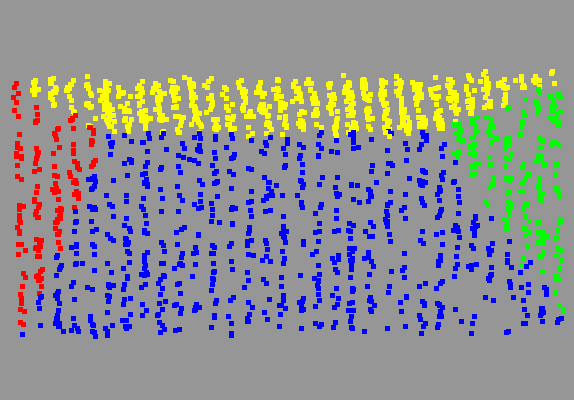}} &
		\multicolumn{1}{c}{\includegraphics[width= 0.225 \linewidth]{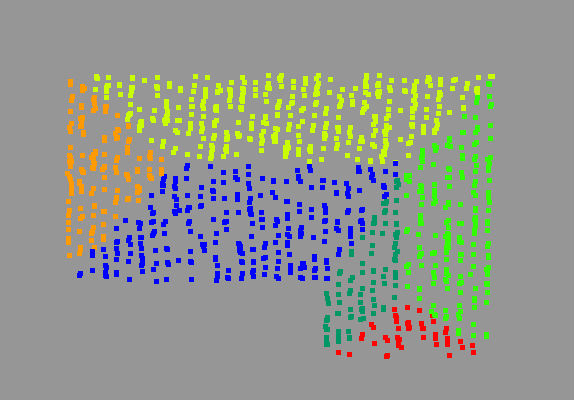}} &
		\multicolumn{1}{c}{\includegraphics[width= 0.225 \linewidth]{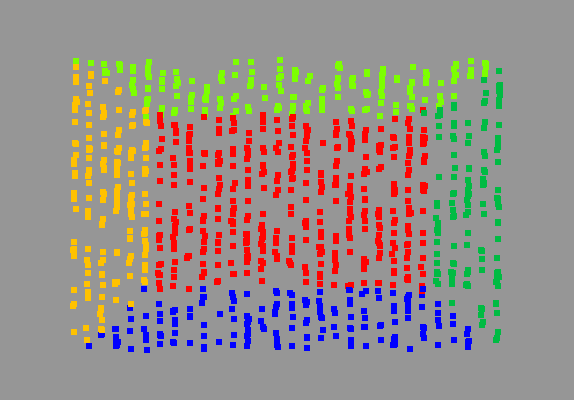}} 	\\
		\multicolumn{1}{c}{\includegraphics[width= 0.225 \linewidth]{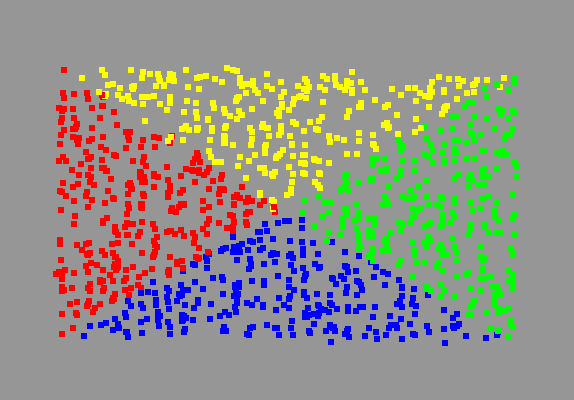}} &
		\multicolumn{1}{c}{\includegraphics[width= 0.225 \linewidth]{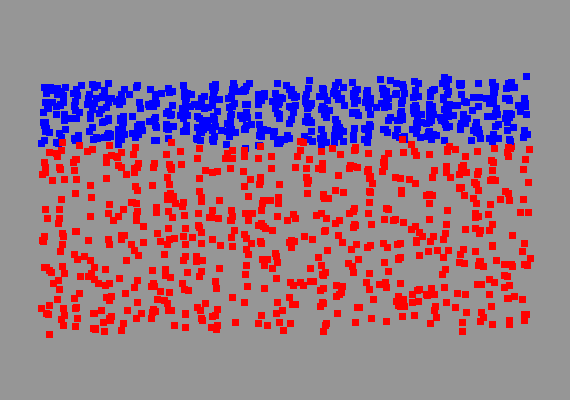}} &
		\multicolumn{1}{c}{\includegraphics[width= 0.225 \linewidth]{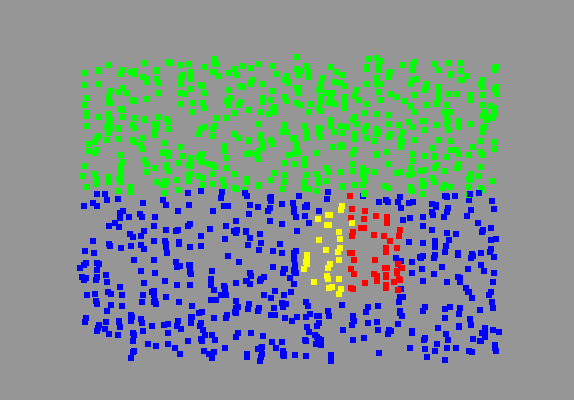}} &
		\multicolumn{1}{c}{\includegraphics[width= 0.225 \linewidth]{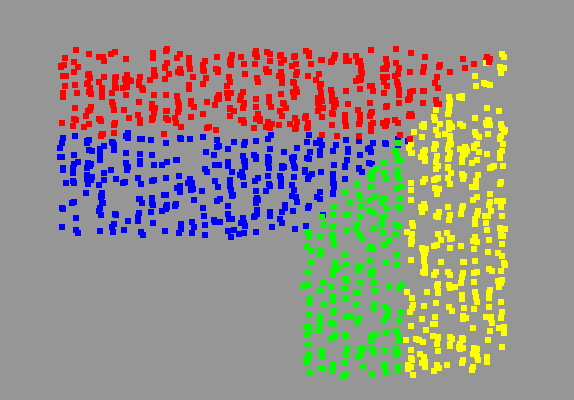}} 	\\
	\end{tabular} 
	\caption{Several examples of the synthetic dataset. Different colors represent different plane instances.}
	\vspace{0.0em}
	\label{Fig:syneticdataset}
\end{figure}

\textbf{Synthetic dataset.} In our previous work Point2Roof~\citep{li2022point2roof}, we construct a synthetic dataset for 3D roof model reconstruction. In this dataset, there are 16 roof types, each roof type includes $1,000$ train buildings and $100$ test buildings. For each building sample, it consists of a roof model and corresponding synthetic airborne LiDAR point clouds. Based on this roof reconstruction dataset, we automatically construct a synthetic dataset for roof plane segmentation. For each building, the points that belong to the same face of the roof model are extracted as a plane instance. We only select $14$ roof types from the roof model reconstruction dataset. This is because the buildings of other $2$ roof types are only comprised of one plane. Thus, we have $15,400$ buildings in our roof plane segmentation dataset. Then, this dataset is split into training and testing sets at a ratio of $10:1$. In Figure~\ref{Fig:syneticdataset}, we presented several examples of our synthetic roof plane segmentation dataset. It should be noted that there are no non-roof points in the synthetic dataset. Thus, for this dataset, we classify the input points into two categories, namely, boundary and plane points.  

\begin{figure}[!htb]
	\centering
	\begin{tabular}{ccc}
		\multicolumn{1}{c}{\includegraphics[width= 0.31 \linewidth]{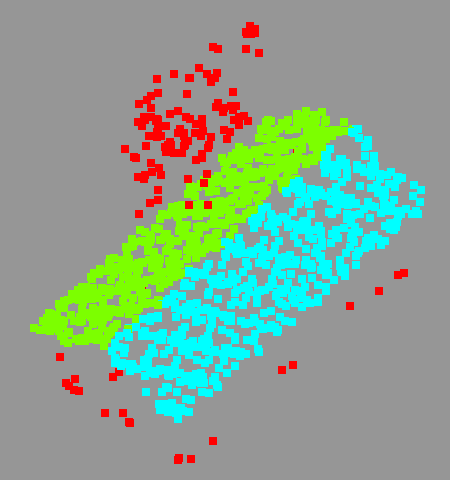}} &
		\multicolumn{1}{c}{\includegraphics[width= 0.31 \linewidth]{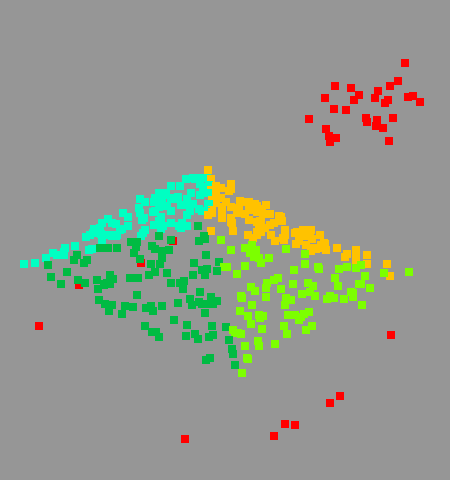}} &
		\multicolumn{1}{c}{\includegraphics[width= 0.31 \linewidth]{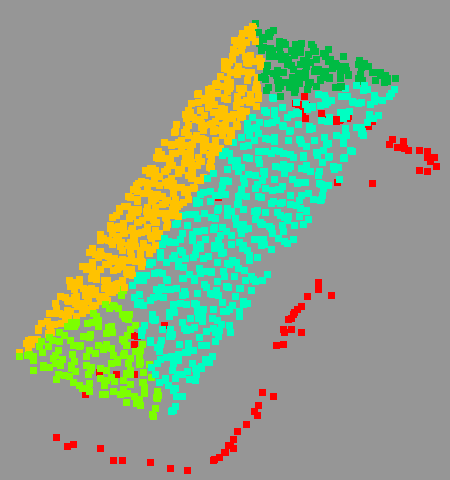}} 	\\
	\end{tabular} 
	\caption{Several examples of RoofN3D dataset. The red dots represent non-roof points, and other different colors represent different plane instances.}
	\vspace{0.0em}
	\label{Fig:realdataset}
\end{figure}

\textbf{RoofN3D dataset.} Except for the synthetic dataset, we also manually construct a real dataset to evaluate the proposed approach. To construct the real dataset, we first select $1,870$ buildings from RoofN3D~\citep{wichmann2019roofn3d}, which is an open 3D building reconstruction dataset. This dataset consists of a large number of real buildings collected from the New York City dataset. In this dataset, there are three roof types: gable, pyramid and hip. For each building, their point clouds are collected by the real airborne LiDAR system. Because the roof planes of each building in RoofN3D are automatically extracted by a traditional region growing-based plane segmentation approach, the ground truth of plane instances may be inaccurate. Thus, we manually assign semantic (non-roof or roof) and instance labels to the selected building points. The real dataset is also split into training and testing sets at a ratio of $10:1$. In Figure~\ref{Fig:realdataset}, we present several samples of the real dataset.

\begin{figure}[!htb]
	\centering
	\footnotesize
	\begin{tabular}{ccc}
		\multicolumn{1}{c}{\includegraphics[width= 0.31 \linewidth]{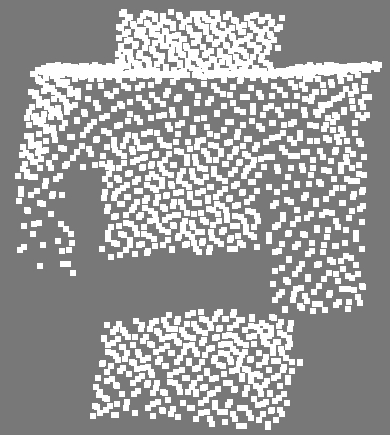}} &
		\multicolumn{1}{c}{\includegraphics[width= 0.31 \linewidth]{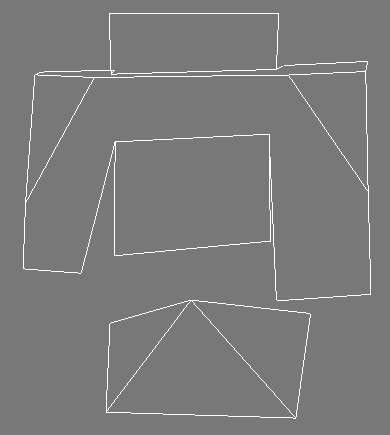}} &
		\multicolumn{1}{c}{\includegraphics[width= 0.31 \linewidth]{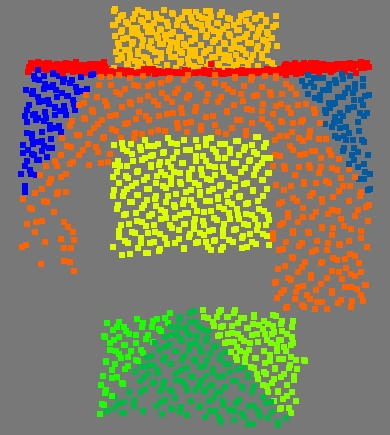}} 	\\
		\multicolumn{3}{c} {(a) An example of ground truth generation. From left to right: input roof points, wireframe model,} \\
	   \multicolumn{3}{c}{ and ground truth of roof planes.}\\
	   \multicolumn{1}{c}{\includegraphics[width= 0.31 \linewidth]{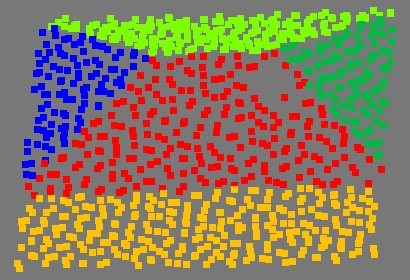}} &
	   \multicolumn{1}{c}{\includegraphics[width= 0.31 \linewidth]{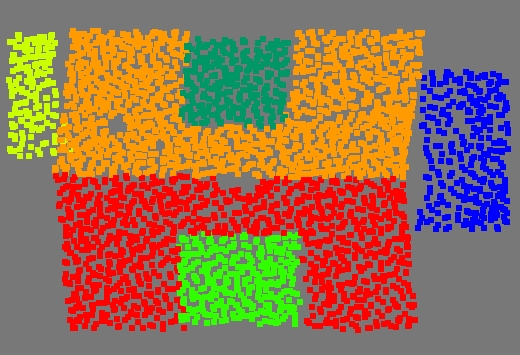}} &
	   \multicolumn{1}{c}{\includegraphics[width= 0.31 \linewidth]{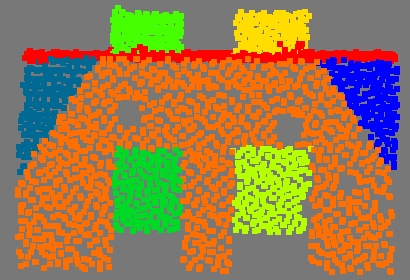}} \\
	   \multicolumn{3}{c} {(b) Several examples of Building3D dataset.}
	\end{tabular} 
	\caption{Visual illustration of Building3D dataset. Different colors represent different plane instances.}
	\vspace{0.0em}
	\label{Fig:Building3D}
\end{figure}

\textbf{Building3D dataset.} Building3D~\citep{wang2023building3d} is an urban-scale building 3D modeling benchmark. This dataset contains a large-scale real building point clouds with corresponding meshes and wireframe models. However, Building3D does not provide ground truth of roof planes. We attempt to directly extract roof planes from wireframe models, as shown in Fig.~\ref{Fig:Building3D}(a). However, we fail to correctly extract roof planes for all buildings because wireframe models do not provide face information. Finally, we obtain $18,227$ buildings with ground truth. We also split these data into training and testing sets at a ratio of $10:1$. This dataset is challenging because there are many buildings with complex roof shapes. In Fig.~\ref{Fig:Building3D}(b), we present several examples of Building3D dataset. Similar to synthetic dataset, there also are no non-roof points in this dataset. 

\subsection{Evaluation metric}
\label{sec:metric}

To quantitatively evaluate the performance of our proposed approach, several evaluation metrics are adopted. In our experiments, coverage (Cov), weighted coverage (WCov), mean precision (mPrec), and mean recall (mRec) metrics are used. mPrec and mRec are two classical evaluation metrics for instance segmentation. For mPrec and mRec, IoU threshold is set to 0.5. Cov is defined as the average instance-wise IoU between prediction and ground truth. To calculate WCov, the value of Cov is further weighted by the number of points included in the corresponding plane instance. The larger values of Cov, WCov, mPrec, and mRec indicate better quality of plane instances. For the detailed descriptions of Cov and WCov, please refer to ASIS~\citep{wang2019associatively}. 

\subsection{Ablation study}
\label{sec:ablation}

There are two key contributions in our approach. The first is that we propose to perform point clustering in both Euclidean and embedding spaces constructed by a deep network. The second is that we identify unreliable boundary points before performing point clustering. For boundary and non-boundary points, we apply different strategies to process them. To convincingly illustrate the effectiveness of these two contributions presented in our approach, we perform ablation studies on all three datasets. 

\begin{table} [!htb] \renewcommand{\tabcolsep}{2.5 pt}
	\scriptsize
	\renewcommand{\arraystretch}{1.2}
	\newcommand{\tabincell}[2]{\begin{tabular}{@{}#1@{}}#2\end{tabular}}
	\begin{center}
		\caption{Quantitative evaluation of our approach with different components on three datasets. Euclid., Emb. and Bound. represent Euclidean space, embedding space and boundary-aware mechanism, respectively. }
		\label{Tab:ablation}
		\begin{tabular}{cccc|cccc|cccc|cccc} 
			\hline
			\multicolumn{4}{c}{Different components} & \multicolumn{4}{c}{Synthetic dataset}  &  \multicolumn{4}{c}{RoofN3D dataset} &  \multicolumn{4}{c}{Building3D dataset}   \\ 
			\cline{1-4} \cline{5-8} \cline{9-12} \cline{13-16} 
			& Euclid.    & Emb.   &   Bound.
			& Cov       & WCov  &   mPrec   &    mRec      
	     	& Cov       & WCov  &   mPrec   &    mRec
	     	& Cov       & WCov  &   mPrec   &    mRec         \\
			\hline 
			\tabincell{c}{(a)}  & \checkmark &   &   & 0.8555  & 0.8995 & 0.9399 & 0.9250 & 0.8408 & 0.8632 & 0.8098 & 0.9532 & 0.8362 & 0.8792 & 0.8058 & 0.8965  \\ 
			\tabincell{c}{(b)}  &  & \checkmark &  & 0.8822  & 0.9292 & 0.9309 & 0.9274 & 0.8556 & 0.8803 & 0.8143 & 0.9579 & 0.8577 & 0.8944 & 0.8353 & 0.9203  \\ 
			\tabincell{c}{(c)}  & \checkmark & \checkmark &  & 0.8994  & 0.9471 & 0.9821 & 0.9321 & 0.8782 & 0.9072 & 0.9497 & 0.9584 & 0.8689 & 0.9055 & 0.8597 & 0.9276   \\ 
			\tabincell{c}{(d)} & \checkmark & \checkmark & \checkmark & 0.9292  & 0.9635 & 0.9976 & 0.9574 & 0.9121 & 0.9161 & 0.9935 & 0.9930 & 0.8875 & 0.9247 & 0.9799 & 0.9322   \\ 
			\hline 
		\end{tabular}
	\end{center}
\end{table}

In Table~\ref{Tab:ablation}, we present the evaluation results of ablation studies. When only Euclidean/embedding space is used, namely, we cut the branch of embedding/Euclidean space construction and we process boundary and non-boundary points together, the scores of metrics are low, as shown in the third and fourth rows of Table~\ref{Tab:ablation}. When both Euclidean and embedding spaces are used, the performance of our approach is significantly improved, as shown in the fifth row of Table~\ref{Tab:ablation}. It should be noted that we still perform unlabeled point refinement for the above three cases.
After using boundary-aware mechanism, the full proposed roof plane segmentation approach achieves the best performance by a large margin, as shown in the last row of Table~\ref{Tab:ablation}. 

In addition, there is a key threshold clustering radius $r$ that needs to be set in our approach. To illustrate the influence of $r$, we report the performance of our approach with the use of different values of $r$ in Table~\ref{Tab:thresholdr}. From this table, we observe that the proposed approach is not sensitive to $r$ in synthetic and RoofN3D datasets. The main reason is that we perform point clustering only for reliable plane points. In these two datasets, the plane points of each instance are close enough in Euclidean and embedding spaces. In addition, the distance between plane points that come from different instances is also large. However, we also find that the proposed method is relatively sensitive to $r$ in Building3D dataset. The main reason is that the buildings included in this dataset are complex. The network may cannot completely distinguish all adjacent plane instances with a large margin in the attribute spaces. This problem can be alleviated by using a more powerful clustering algorithm. In our experiments, we empirically set $r = 0.1$ for three datasets. 

\begin{table} [!htb] \renewcommand{\tabcolsep}{3.5 pt}
	\scriptsize
	\renewcommand{\arraystretch}{1.2}
	\newcommand{\tabincell}[2]{\begin{tabular}{@{}#1@{}}#2\end{tabular}}
	\begin{center}
		\caption{Ablation results of our approach with different values of $r$ on all three datasets. }
		\label{Tab:thresholdr}
		\begin{tabular}{c|cccc|cccc|cccc} 
			\hline
			\multirow{2}*{Values of $r$} & \multicolumn{4}{c}{Synthetic dataset}  &  \multicolumn{4}{c}{RoofN3D dataset} & \multicolumn{4}{c}{Building3D dataset}  \\ 
			\cline{2-5} \cline{6-9} \cline{10-13} 
			& Cov       & WCov  &   mPrec   &    mRec      
			& Cov       & WCov  &   mPrec   &    mRec
			& Cov       & WCov  &   mPrec   &    mRec         \\
			\hline 
			$r = 0.05$ & 0.9246 & 0.9611 & 0.9946 & 0.9531 & 0.9090 & 0.9139 & 0.9875 & 0.9875 & 0.8381 & 0.8789 & 0.9053 & 0.8853  \\ 
			$r = 0.10$   & 0.9292 & 0.9635 & 0.9976 & 0.9574 & 0.9121 & 0.9161 & 0.9935 & 0.9930 & 0.8875 & 0.9247 & 0.9799 & 0.9322  \\ 
			$r = 0.15$ & 0.9315 & 0.9643 & 0.9983 & 0.9596 & 0.9094 & 0.9138 & 0.9926 & 0.9887 & 0.8705 & 0.9117 & 0.9786 & 0.9049   \\ 
			$r = 0.20$   & 0.9319 & 0.9636 & 0.9989 & 0.9599 & 0.9079 & 0.9129 & 0.9866 & 0.9881 & 0.8485 & 0.8942 & 0.9732 & 0.8751   \\ 
			$r = 0.25$   & 0.9303  & 0.9617 & 0.9993 & 0.9579 & 0.8773 &  0.9095 & 0.9910 & 0.9542 & 0.8246 & 0.8751 & 0.9641 & 0.8436    \\ 
			\hline 
		\end{tabular}
	\end{center}
\end{table}

\subsection{Comparative experiments}
\label{sec:compare}

We compared the proposed roof plane segmentation approaches with RANSAC~\citep{schnabel2007efficient}, region growing~\citep{lafarge2012creating}, PointGroup~\citep{jiang2020pointgroup}, Zhu et al.'s approach~\citep{zhu2021robust}, GoCoPP~\citep{yu2022finding} and Liu et al.'s approach~\citep{liu2023roof} using $1,400$ synthetic, $170$ RoofN3D and $1,657$ Building3D test buildings. PointGroup is a deep network designed for 3D point cloud instance segmentation, and the rest are traditional plane segmentation approaches. For RANSAC and region growing, their implementations are available at CGAL~\footnote{Available at \url{https://www.cgal.org/}.}, which is an open library. The source codes of PointGroup and Liu et al.'s approach are provided by the authors. The source code of GoCoPP is downloaded from homepage~\footnote{Website: \url{https://team.inria.fr/titane/}} of the team that the authors belong to. For Zhu et al.'s approach, it has been implemented by a commercial software~\emph{LiDARPro}~\footnote{Available at \url{https://skyearth.org/LiDARPro/}.}. This software has been widely applied for 3D building model reconstruction in many real applications.

In Table~\ref{Tab:compara}, we report the quantitative evaluation results of all approaches on synthetic and two real datasets. From Table~\ref{Tab:compara}, it is obvious that our approach offers the best scores of all metrics on three datasets. In addition, the scores of our approach are significantly higher than the scores produced by the other approaches. For synthetic dataset, the scores of Cov and WCov generated by our approach are $0.9292$ and $0.9635$. For the rest approaches, PointGroup offers the best scores of Cov ($0.8308$) and WCov ($0.8705$), which are obviously smaller than ours. It means that the proposed approach can accurately predict the plane instance label for each point. In addition, our approach also produces the highest scores of mPrec ($0.9976$) and mRec ($0.9574$) on synthetic dataset, this implies that our approach can accurately extract planes at instance level. For RoofN3D and Building3D datasets, it is easy to get the similar observations and conclusions from Table~\ref{Tab:compara}.

\begin{table} [!htb] \renewcommand{\tabcolsep}{2 pt}
	\scriptsize
	\renewcommand{\arraystretch}{1.2}
	\newcommand{\tabincell}[2]{\begin{tabular}{@{}#1@{}}#2\end{tabular}}
	\begin{center}
		\caption{Quantitative evaluation of all approaches on three datasets. }
		\label{Tab:compara}
		\begin{tabular}{c|cccc|cccc|cccc} 
			\hline
			\multirow{2}*{Different approaches} & \multicolumn{4}{c}{Synthetic dataset}  &  \multicolumn{4}{c}{RoofN3D dataset} & \multicolumn{4}{c}{Building3D dataset}  \\ 
			 \cline{2-5} \cline{6-9} \cline{10-13}
			& Cov       & WCov  &   mPrec   &    mRec      
			& Cov       & WCov  &   mPrec   &    mRec
			& Cov       & WCov  &   mPrec   &    mRec         \\
			\hline 
			 RANSAC & 0.6795  & 0.7300 & 0.6606 & 0.8182 & 0.7103 & 0.7756 & 0.5368 & 0.7866 & 0.7295 & 0.7608 & 0.6679 & 0.7985  \\ 
			 Region growing   & 0.6523  & 0.7125 & 0.6013 & 0.7651 & 0.7123 & 0.7792 & 0.6423 & 0.8109 & 0.7054 & 0.7250 & 0.5789 & 0.7532  \\ 
			 PointGroup & 0.8308  & 0.8705 & 0.7662 & 0.9353 & 0.7758  & 0.7952 & 0.9107 & 0.8783 & 0.7215 & 0.7630 & 0.6858 & 0.7602   \\ 
			 Zhu et al.'s approach   & 0.8111  & 0.8570 & 0.8475 & 0.8723 & 0.6978 & 0.7738 & 0.6965 & 0.7897 & 0.7189 & 0.7606 & 0.7938 & 0.7624    \\
			 GoCoPP &  0.8065 & 0.8283 & 0.7521 & 0.9083 & 0.7227 & 0.7256 & 0.6405 & 0.7831 & 0.7820 & 0.7973 & 0.7367 & 0.8441  \\
			 Liu et al.'s approach   & 0.8127  & 0.8499 & 0.9003 & 0.8785 & 0.7446 & 0.8264 & 0.8565 & 0.8477 & 0.7557 & 0.7894 & 0.7577 & 0.8063    \\ 
			 Our approach & \textbf{0.9292}  & \textbf{0.9635} & \textbf{0.9976} & \textbf{0.9574} & \textbf{0.9121} & \textbf{0.9161} & \textbf{0.9935} & \textbf{0.9930} & \textbf{0.8875} & \textbf{0.9247} & \textbf{0.9799} & \textbf{0.9322}   \\ 
			\hline 
		\end{tabular}
	\end{center}
\end{table}

\begin{figure}[!htb]
	\scriptsize
	\centering
\begin{tabular}{ccc} 	
	\multicolumn{1}{c}{\includegraphics[width= 0.31 \linewidth]{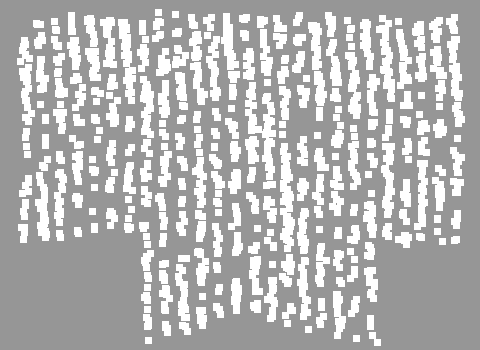}} &
	\multicolumn{1}{c}{\includegraphics[width= 0.31 \linewidth]{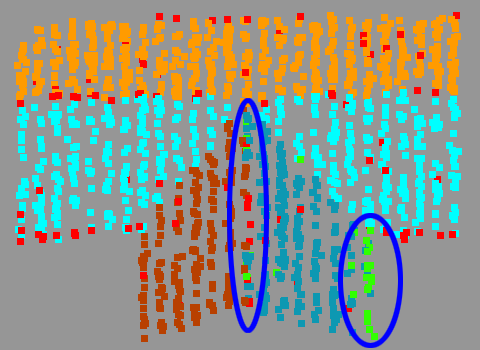}} &
	\multicolumn{1}{c}{\includegraphics[width= 0.31 \linewidth]{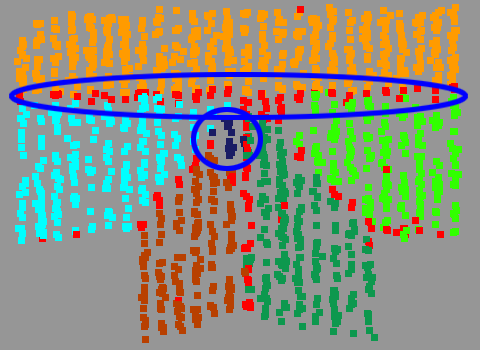}} \\
	(a) Input & (b) RANSAC & (c) Region growing                 \\
	
	\multicolumn{1}{c}{\includegraphics[width= 0.31 \linewidth]{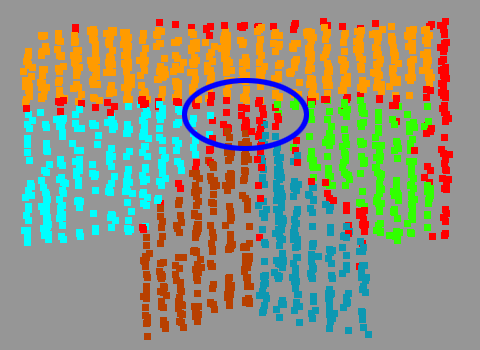}} &
	\multicolumn{1}{c}{\includegraphics[width= 0.31 \linewidth]{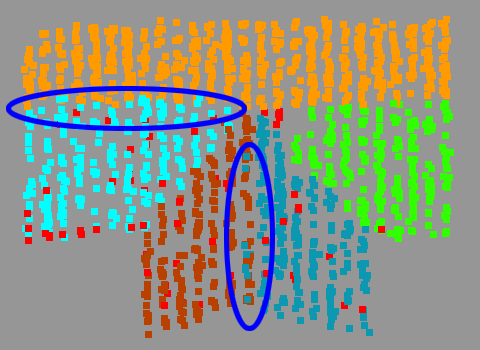}} &
	\multicolumn{1}{c}{\includegraphics[width= 0.31 \linewidth]{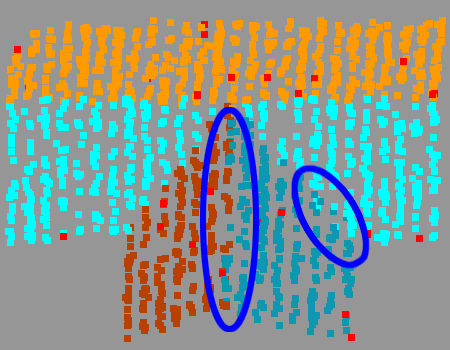}}  \\
	(d) PointGroup & (e)  \cite{zhu2021robust}  &  (f) GoCoPP  \\
	
	\multicolumn{1}{c}{\includegraphics[width= 0.31 \linewidth]{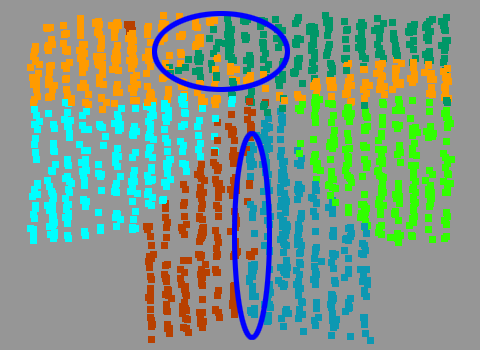}} &
	\multicolumn{1}{c}{\includegraphics[width= 0.31	\linewidth]{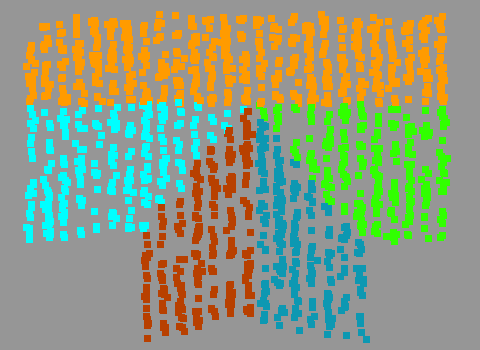}} &
	\multicolumn{1}{c}{\includegraphics[width= 0.31 \linewidth]{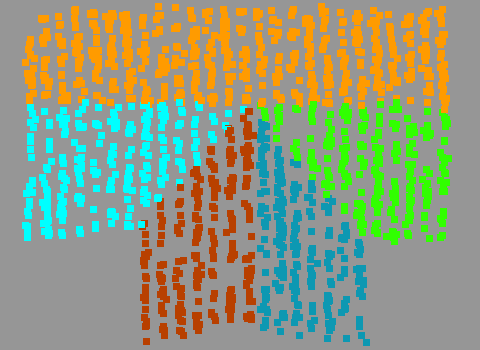}} 	\\
	(g) \cite{liu2023roof} & (h) Our approach  & (i) Ground truth  \\
\end{tabular} 
	\caption{A visual example of plane instance segmentation results of different approaches on the synthetic dataset. The red points represent unlabeled points and other different colors represent different plane instances. The blue ellipses are applied to highlight some representative problem areas.}
	\vspace{0.0em}
	\label{Fig:Compaer1}
\end{figure}

\begin{figure}[!htb]
	\scriptsize
	\centering
\begin{tabular}{ccc} 	
		\multicolumn{1}{c}{\includegraphics[width= 0.25 \linewidth]{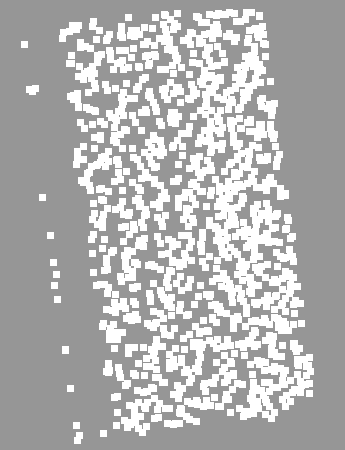}} &
		\multicolumn{1}{c}{\includegraphics[width= 0.25 \linewidth]{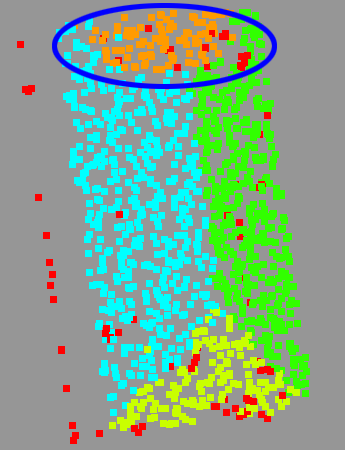}} &
		\multicolumn{1}{c}{\includegraphics[width= 0.25 \linewidth]{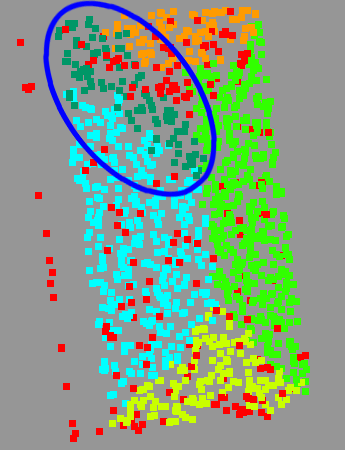}} \\
		(a) Input & (b) RANSAC & (c) Region growing                 \\
		
		\multicolumn{1}{c}{\includegraphics[width= 0.25 \linewidth]{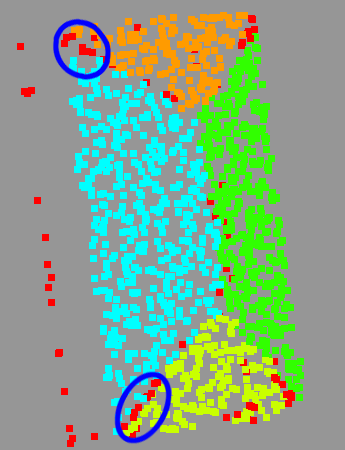}} &
		\multicolumn{1}{c}{\includegraphics[width= 0.25 \linewidth]{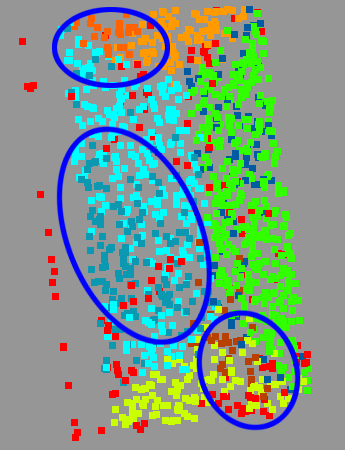}} &
		\multicolumn{1}{c}{\includegraphics[width= 0.25 \linewidth]{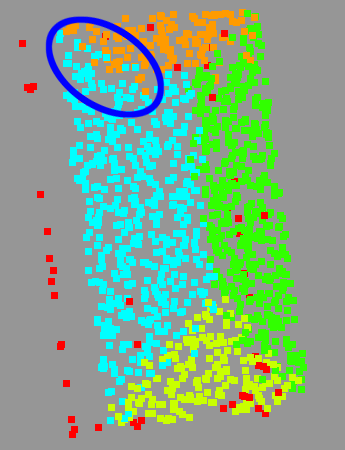}}  \\
		(d) PointGroup & (e)  \cite{zhu2021robust}  &  (f) GoCoPP  \\
		
		\multicolumn{1}{c}{\includegraphics[width= 0.25 \linewidth]{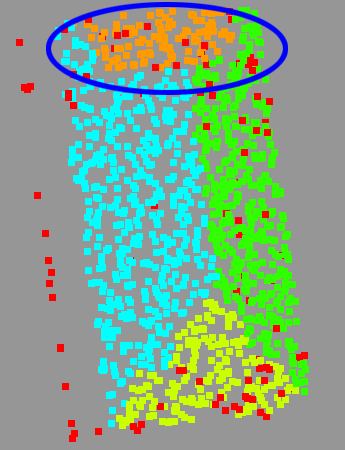}} &
		\multicolumn{1}{c}{\includegraphics[width= 0.25	\linewidth]{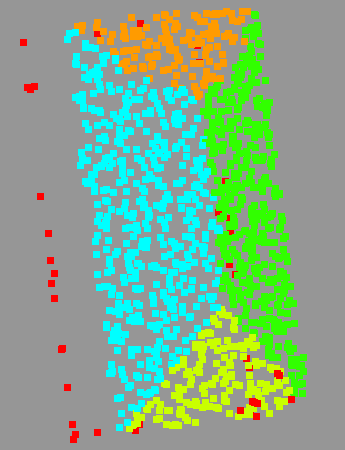}} &
		\multicolumn{1}{c}{\includegraphics[width= 0.25 \linewidth]{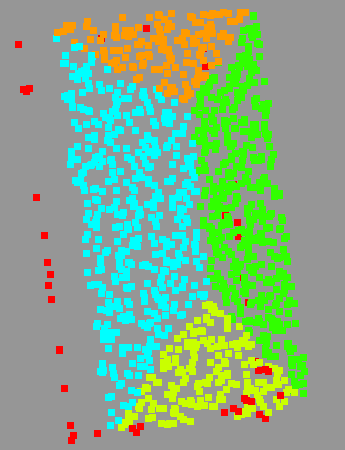}} 	\\
		(g) \cite{liu2023roof} & (h) Our approach  & (i) Ground truth  \\
\end{tabular} 
	\caption{A visual example of plane instance segmentation results of different approaches on the RoofN3D dataset. The red points represent unlabeled points and other different colors represent different plane instances. The blue ellipses are applied to highlight  some representative problem areas.}
	\vspace{0.0em}
	\label{Fig:Compaer2}
\end{figure}

\begin{figure}[!htb]
	\scriptsize
	\centering
\begin{tabular}{ccc} 	
		\multicolumn{1}{c}{\includegraphics[width= 0.30 \linewidth]{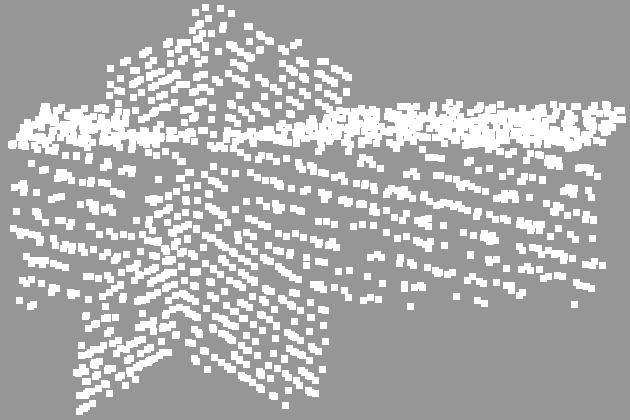}} &
		\multicolumn{1}{c}{\includegraphics[width= 0.30 \linewidth]{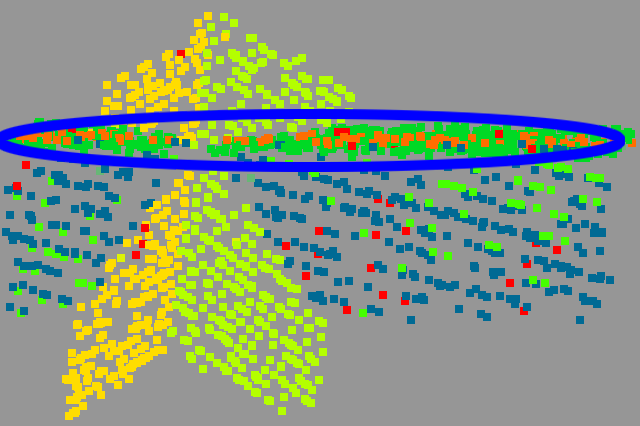}} &
		\multicolumn{1}{c}{\includegraphics[width= 0.30 \linewidth]{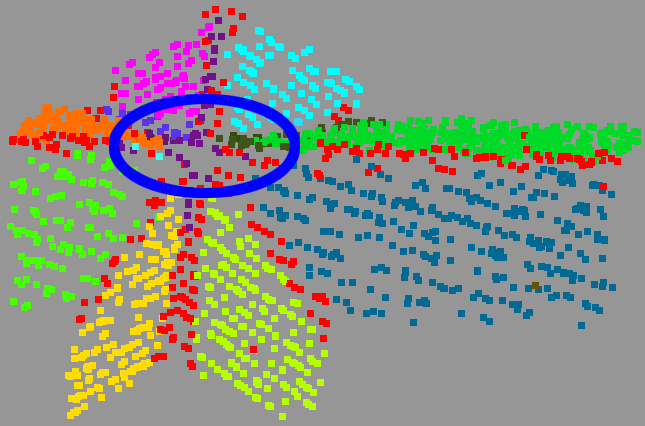}} \\
		(a) Input & (b) RANSAC & (c) Region growing                 \\
		
		\multicolumn{1}{c}{\includegraphics[width= 0.30 \linewidth]{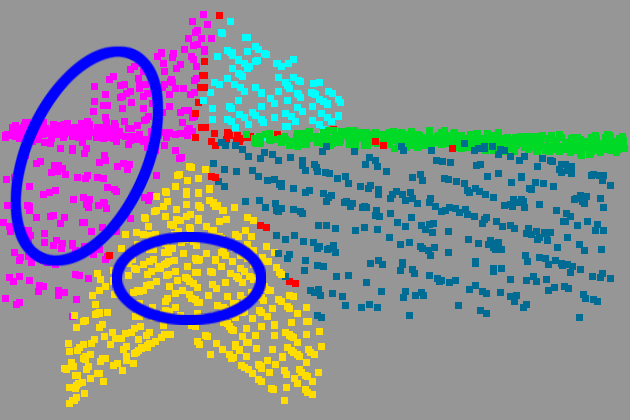}} &
		\multicolumn{1}{c}{\includegraphics[width= 0.30 \linewidth]{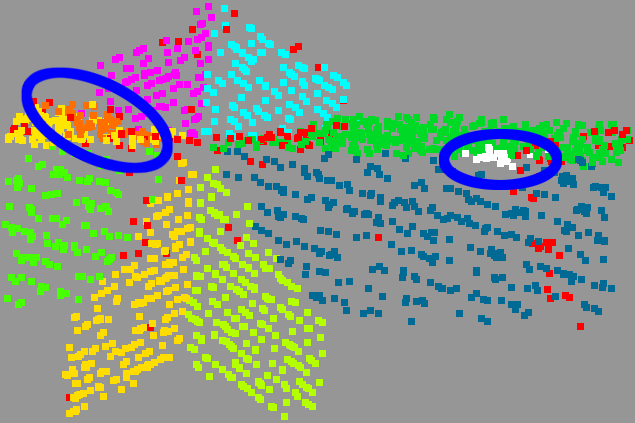}} &
		\multicolumn{1}{c}{\includegraphics[width= 0.30 \linewidth]{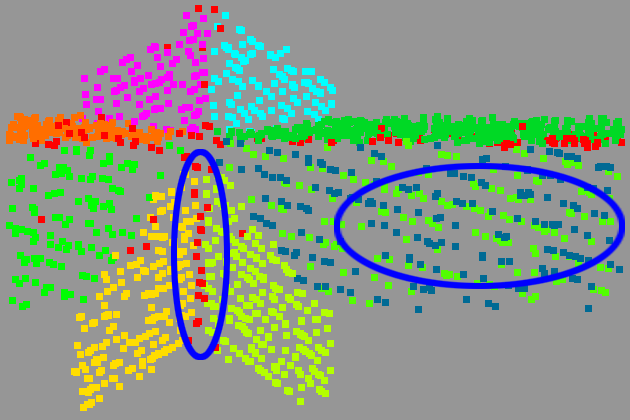}}  \\
		(d) PointGroup & (e)  \cite{zhu2021robust}  &  (f) GoCoPP  \\
		
		\multicolumn{1}{c}{\includegraphics[width= 0.30 \linewidth]{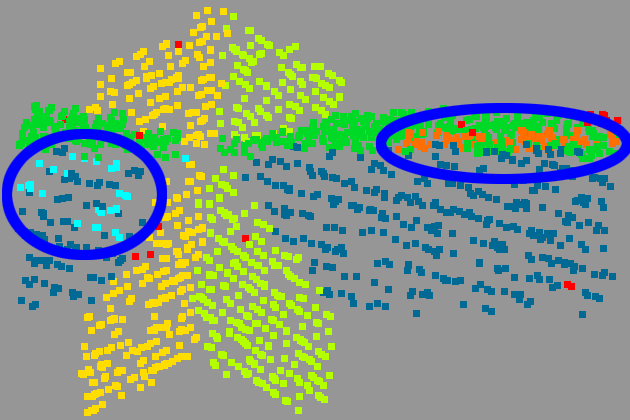}} &
		\multicolumn{1}{c}{\includegraphics[width= 0.30	\linewidth]{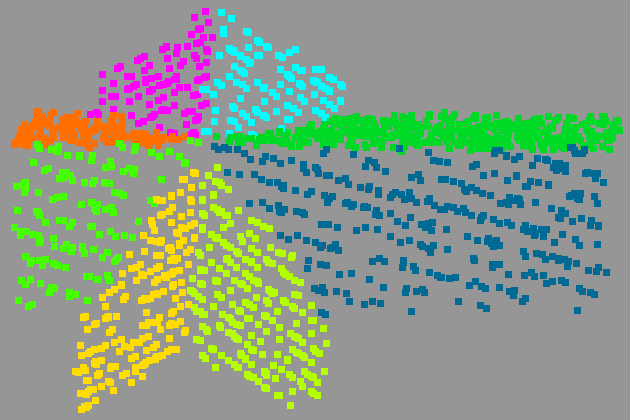}} &
		\multicolumn{1}{c}{\includegraphics[width= 0.30 \linewidth]{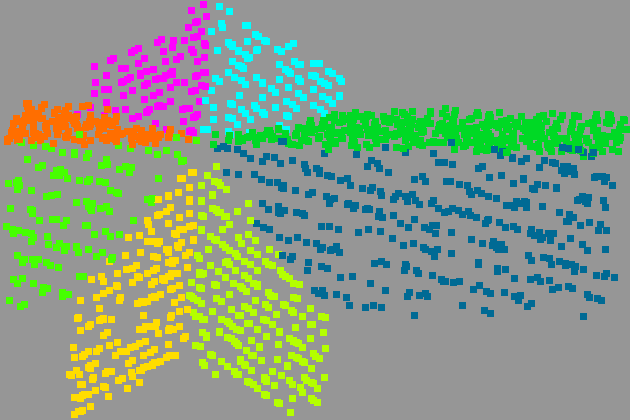}} 	\\
		(g) \cite{liu2023roof} & (h) Our approach  & (i) Ground truth  \\
\end{tabular} 
	\caption{A visual example of plane instance segmentation results of different approaches on the Building3D dataset. The red points represent unlabeled points and other different colors represent different plane instances. The blue ellipses are applied to highlight  some representative problem areas.}
	\vspace{0.0em}
	\label{Fig:Compaer3}
\end{figure}

In addition to quantitative evaluation, to visually illustrate that the proposed method can generate high-quality roof planes, we also present some roof plane extraction results of all approaches in Figure~\ref{Fig:Compaer1}, Figure~\ref{Fig:Compaer2} and Figure~\ref{Fig:Compaer3}. In Figure~\ref{Fig:Compaer1}, we visually present plane segmentation results of different approaches on a building selected from the synthetic test dataset. In Figure~\ref{Fig:Compaer1}(b), the plane segmentation result of RANSAC is presented. RANSAC produces a small spurious plane and inaccurate boundaries. In addition, RANSAC merges two plane instances into one large plane. Region growing generates a small spurious plane and fails to find the correct plane instances for boundary points, as shown in Figure~\ref{Fig:Compaer1}(c). PointGroup successfully extracts five plane instances from the input data, but it struggles to process unreliable boundary points, as shown in Figure~\ref{Fig:Compaer1}(d). PointGroup fails to assign the correct instance labels to the boundary points. Zhu et al.'s approach and GoCoPP generate inaccurate boundaries between adjacent plane instances, as highlighted by the blue ellipses in Figure~\ref{Fig:Compaer1}(e) and (f). In addition, GoCoPP merges two plane instances into one plane. Liu et al.'s approach suffers from over-segmentation problem, as shown in Figure~\ref{Fig:Compaer1}(g). A roof plane has been split into two planes. In addition, this approach also generates inaccurate boundaries. The proposed approach produces the best result for this data, the segmented planes are almost the same with ground truth, as shown in Figure~\ref{Fig:Compaer1}(h). 

\begin{figure}[!htb]
	\scriptsize
	\centering
\begin{tabular}{cccc} 	
	\multicolumn{4}{c}{\includegraphics[width= 0.99 \linewidth]{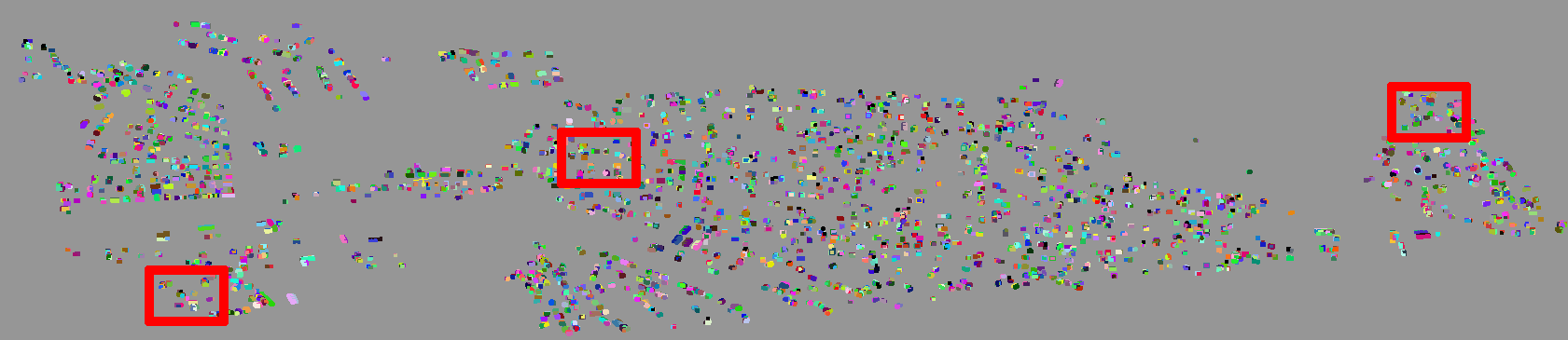}}  \vspace{0.5 em} \\ 
	\multicolumn{1}{c}{\includegraphics[width= 0.312  \linewidth]{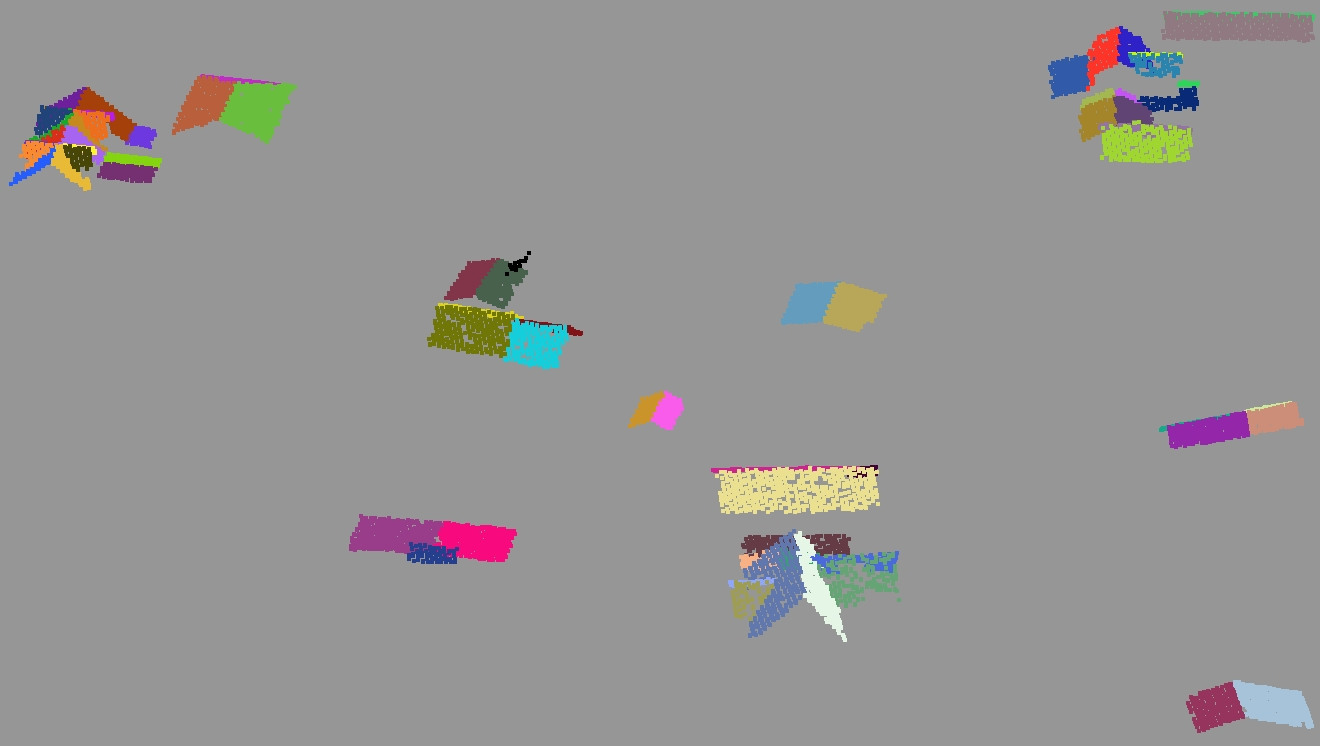}} &
	\multicolumn{1}{c}{\includegraphics[width= 0.312 \linewidth]{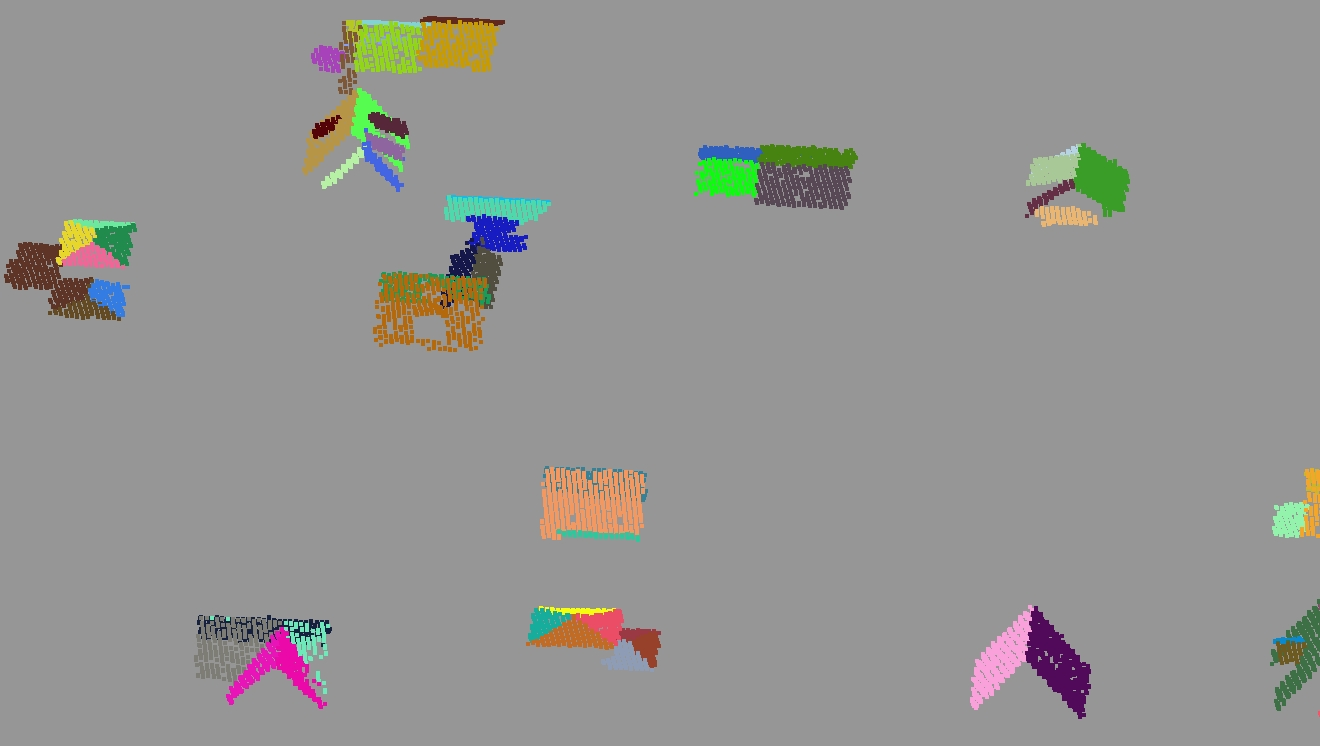}} &
	\multicolumn{1}{c}{\includegraphics[width= 0.312 \linewidth]{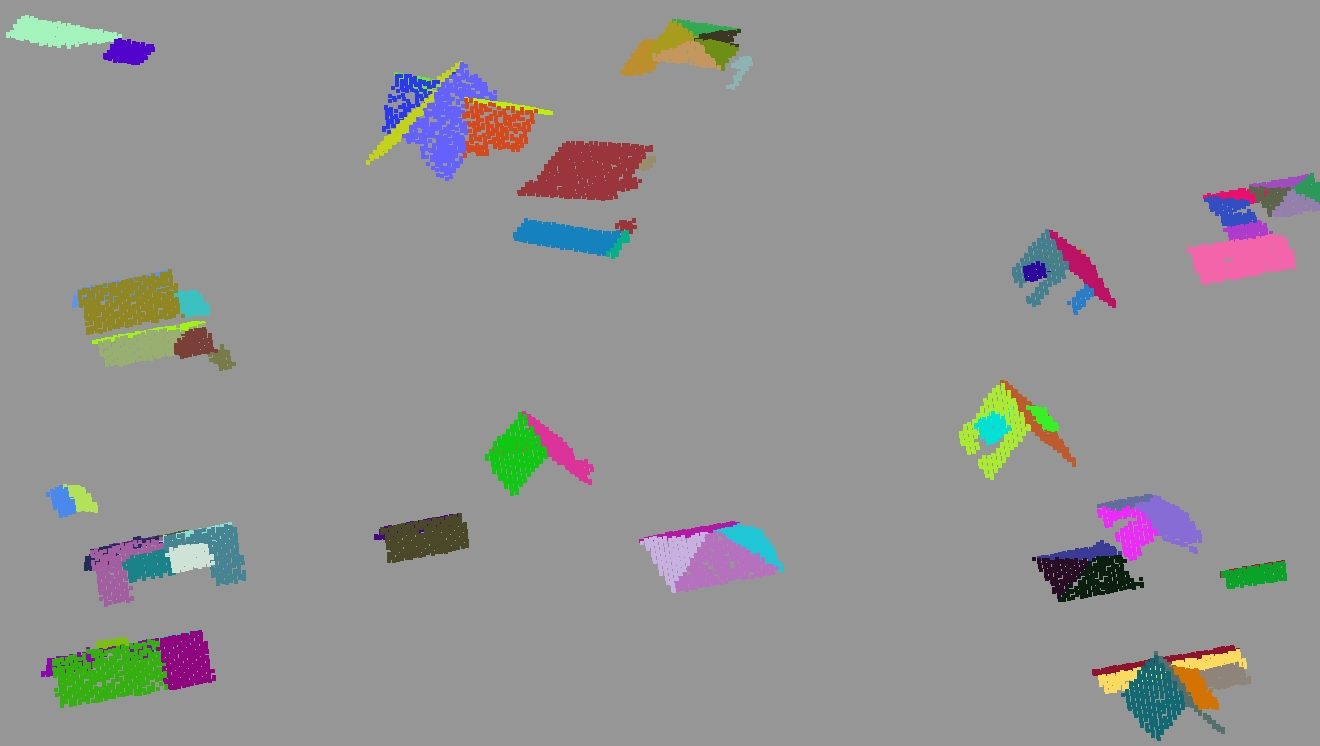}} \\
\end{tabular} 
	\caption{Plane segmentation results of our approach on the whole Building3D test set. Three enlarged regions highlighted by the red rectangles are presented in the second row. Different colors represent different plane instances.}
	\vspace{0.0em}
	\label{Fig:largeScene}
\end{figure}

In Figure~\ref{Fig:Compaer2}, the plane segmentation results of a real building selected from RoofN3D dataset are presented. We observe that the result generated by our approach is the best. RANSAC and Liu et al.'s approach produce inaccurate boundaries. Both region growing and Zhu et al.'s approach generate spurious planes and inaccurate boundaries. PointGroup performs better in this data, but it still fails to assign the correct instance labels to some boundary points. GoCoPP performs well in this data, but it still fails to generate smooth and accurate boundaries. In Figure~\ref{Fig:Compaer3}, we present the results of a building selected from Building3D dataset. The proposed approach also generates the best result. RANSAC, Region growing, Zhu et al.'s and Liu et al.'s approaches generate spurious planes. Region growing also struggles to process boundary points. PointGroup suffers from severe under-segmentation problem. The main reason is that PointGroup fails to completely distinguish tightly connected roof planes. GoCoPP fails to assign the correct instance labels to some boundary points, and detects two nearly
parallel and overlapped plane instances in the region highlighted by the larger blue ellipse, as shown in Figure~\ref{Fig:Compaer3}(f). In addition, in Figure~\ref{Fig:largeScene}, we visually present the roof plane segmentation results of the proposed approach on the whole Building3D test set. We also make the segmentation results presented in Figure~\ref{Fig:largeScene} publicly available at \url{https://github.com/Li-Li-Whu/DeepRoofPlane/tree/master/Building3D_Result}. The readers can download our data and open it using \emph{CloudCompare}~\footnote{Available at \url{http://www.cloudcompare.org/}.}.

\subsection{Computational complexity analysis}
\label{sec:cost}

In our experiments, the deep network is implemented with the use of PyTorch and is tested on a single NVIDIA RTX 3090 GPU. Point clustering and refinement algorithms are implemented with C++ under Windows and are tested on a computer with an AMD Ryzen7 6800H at 3.2GHz. The FLOPS and parameters of our network are 0.97G and 1.81M, respectively. 

In Table~\ref{Tab:computation}, we report the running time and GPU consumption of the proposed approach with different numbers of input points. As the number of points increases, the network inference time and GPU memory consumption slightly increase, and the point clustering time also significantly increases. From Table~\ref{Tab:computation}, we observe that the point clustering and network inference spend most of time in our approach, and the running time of point refinement algorithm is very low.

\begin{table} [!htb] \renewcommand{\tabcolsep}{4 pt}
	\scriptsize
	\renewcommand{\arraystretch}{1.2}
	\newcommand{\tabincell}[2]{\begin{tabular}{@{}#1@{}}#2\end{tabular}}
	\begin{center}
		\caption{Running time and GPU memory consumption of the proposed approach.}
		\label{Tab:computation}
		\begin{tabular}{c|ccccc} 
			\hline
			Point numbers & Network inference (s)&  GPU memory (M) & Point clustering (s) & Point refinement (ms) & Total time (s) \\
			\hline
			1024 & 0.72 & 95.12  & 0.04 & 0.23 & 0.76 \\ 
			\hline
			2048 & 0.77 & 102.39 & 0.08 & 0.87 & 0.85 \\ 
			\hline
			4096 & 0.78 & 116.94 & 0.20 & 2.81 & 0.98 \\ 
			\hline
			8192 & 0.79 & 146.03 & 0.58 & 5.24 & 1.38 \\ 
			\hline
		\end{tabular}
	\end{center}
\end{table}

\section{Conclusion}
\label{sec:conclusion}

In this paper, a boundary-aware point clustering approach in Euclidean and embedding spaces constructed by the deep network is proposed to accurately segment roof planes from airborne LiDAR points. Instead of using handcrafted features to distinguish different plane instances, we design a three-branch and multi-task network to extract powerful point features that distinguish plane instances in both Euclidean and embedding spaces. We attempt to ensure that points of the same plane instance are close enough in both spaces. In addition, to avoid the influence of unreliable boundary points, we identify them before performing point clustering and process them using a simple but effective boundary refinement method. The experimental results on synthetic and real datasets illustrate that the proposed roof plane segmentation approach can accurately segment roof planes, and significantly outperforms traditional and deep learning-based approaches.

Nevertheless, the proposed roof plane segmentation approach can be further improved in the following ways. First, the proposed approach is not end-to-end, the accumulated errors may appear and the final extracted roof planes may be not optimal. Thus, we attempt to design an end-to-end network to segment roof planes in the future work. Second, the network architecture is very simple and point features learned by the network may be not powerful enough. We suggest to re-design the network using some advanced modules, such as transformer and deformable attention. Third, there still are several parameters that need to be set in the proposed approach. We suggest to design a parameter-free point clustering method for roof plane segmentation.


\section*{Acknowledgements}

This work was partially supported by the National Natural Science Foundation of China (No.U22A2009, No.42271445), the Fundamental Research Funds for the Central Universities (2042023kf0174), the Foundation of Anhui Province Key Laboratory of Physical Geographic Environment (2022PGE008), the Shenzhen Science and Technology Program (JCYJ20230807090206013), the Guangdong Basic and Applied Basic Research Foundation (2024A1515010218).

\bibliographystyle{elsarticle-harv}
\bibliography{../Refs/refs}

\end{document}